\documentclass[runningheads,final]{llncs}
\usepackage[T1]{fontenc}
\usepackage{graphicx}

\usepackage{algorithm}
\usepackage[noend]{algpseudocode}
\usepackage{amsmath}
\usepackage{amssymb}
\usepackage{array}
\usepackage{cite}
\usepackage{enumitem}
\usepackage{graphicx}
\usepackage{makecell}
\usepackage{multirow}
\usepackage{siunitx}
\usepackage{wasysym}

\usepackage{graphicx}

\providecommand{\rmd}{\mathrm{d}}

\providecommand{\E}{\mathbb{E}}

\providecommand{\R}{\mathbb{R}}

\providecommand{\Spp}{\mathbb{S}_{++}}

\providecommand{\fI}{f_{\mathrm{I}}}
\providecommand{\fII}{f_{\mathrm{II}}}
\providecommand{\fIII}{f_{\mathrm{III}}}

\providecommand{\fIpIV}{f_{\mathrm{I}+\mathrm{IV}}}
\providecommand{\fIIpIV}{f_{\mathrm{II}+\mathrm{IV}}}
\providecommand{\fIIIpIV}{f_{\mathrm{III}+\mathrm{IV}}}

\providecommand{\mcc}{\mathcal{C}}
\providecommand{\mcn}{\mathcal{N}}
\providecommand{\mcu}{\mathcal{U}}
\providecommand{\mcx}{\mathcal{X}}

\providecommand{\argmin}{\operatornamewithlimits{arg\,min}}

\providecommand{\diag}{\operatorname{diag}}

\providecommand{\vecop}{\operatorname{vec}}

\providecommand{\Aa}{\boldsymbol{A}}

\providecommand{\DD}{\boldsymbol{D}}
\providecommand{\FF}{\boldsymbol{F}}

\providecommand{\HH}{\boldsymbol{H}}
\providecommand{\II}{\boldsymbol{I}}
\providecommand{\MM}{\boldsymbol{M}}
\providecommand{\Mh}{\boldsymbol{\hat{M}}}
\providecommand{\QQ}{\boldsymbol{Q}}

\providecommand{\Zero}{\boldsymbol{0}}
\providecommand{\One}{\boldsymbol{1}}

\providecommand{\Cat}{\operatorname{Cat}}

\providecommand{\om}{\operatorname{OneMax}}

\providecommand{\gq}{\boldsymbol{G}_{\boldsymbol{q}}}
\providecommand{\ny}{{N_{\omega}}}

\providecommand{\cc}{\boldsymbol{c}}
\providecommand{\bs}{\boldsymbol{b^\star}}
\providecommand{\bsh}{\boldsymbol{\hat{b}^\star}}

\providecommand{\cs}{\boldsymbol{c^\star}}
\providecommand{\Vs}{\boldsymbol{V^\star}}
\providecommand{\Vsh}{\boldsymbol{\hat{V}^\star}}
\providecommand{\pphis}{\boldsymbol{\phi^\star}}

\providecommand{\mm}{\boldsymbol{m}}
\providecommand{\mt}{\boldsymbol{\tilde m}}

\providecommand{\qq}{\boldsymbol{q}}
\providecommand{\sS}{\boldsymbol{s}}

\providecommand{\xx}{\boldsymbol{x}}
\providecommand{\xs}{\boldsymbol{x^\star}}
\providecommand{\xh}{\boldsymbol{\hat{x}}}
\providecommand{\xhp}{\boldsymbol{\hat{x}}^{\boldsymbol{\prime}}}

\providecommand{\SSigma}{\boldsymbol{\Sigma}}
\providecommand{\SSigmat}{\boldsymbol{\tilde \Sigma}}
\providecommand{\LLambda}{{\boldsymbol{\Lambda}}}
\providecommand{\T}{^{\top}}
\providecommand{\Tmin}{T_{\mathrm{min}}}
\providecommand{\ind}[1]{\mathbb{I}\{#1\}}

\providecommand{\oomega}{\boldsymbol{\omega}}
\providecommand{\ttheta}{\boldsymbol{\theta}}

\providecommand{\oomegat}{\boldsymbol{\tilde{\omega}}}

\providecommand{\pthr}{p_{\mathrm{threshold}}}
\providecommand{\pp}{\bar{p}_{+}}
\providecommand{\pn}{\bar{p}_{-}}

\providecommand{\Vxmin}{V_{\mathrm{min}}}
\providecommand{\Cxmax}{\kappa_{\mathrm{max}}}

\usepackage{xspace}
\providecommand{\proposedmethod}{information-geometric bilevel
decomposition\xspace}
\providecommand{\prme}{IGBD\xspace}

\usepackage{hyperref,cleveref}

\crefformat{equation}{(#2#1#3)}
\Crefformat{equation}{(#2#1#3)}
\crefrangeformat{equation}{(#3#1#4)--(#5#2#6)}
\Crefrangeformat{equation}{(#3#1#4)--(#5#2#6)}
\crefmultiformat{equation}{(#2#1#3)}{ and~(#2#1#3)}{, (#2#1#3)}{, and~(#2#1#3)}
\Crefmultiformat{equation}{(#2#1#3)}{ and~(#2#1#3)}{, (#2#1#3)}{, and~(#2#1#3)}

\usepackage{hyperref}
\usepackage{color}
\begin{document}
\title{Mixed-Categorical Black-Box Optimization via
Information-Geometric Bilevel Decomposition}
\titlerunning{Mixed Optimization via Information-Geometric Bilevel
Decomposition}

\author{Marc Ong\inst{1} \and Shinichi Shirakawa\inst{2} \and Youhei
Akimoto\inst{1,3}}
\institute{
  University of Tsukuba, Tsukuba, Japan \and
  Yokohama National University, Yokohama, Japan \and
  RIKEN Center for Advanced Intelligence Project, Tokyo, Japan
  \email{
    marc@bbo.cs.tsukuba.ac.jp,
    shirakawa-shinichi-bg@ynu.ac.jp,
    akimoto@cs.tsukuba.ac.jp
  }
}
\maketitle              %
\begin{abstract} Mixed categorical-continuous optimization arises in
  many practical domains,
  yet remains challenging. In the black-box setting, evolution
  strategy-based approaches have shown promise in extending the efficiency and
  robustness of the CMA-ES to mixed-variable spaces. However, these methods
  exhibit worsened performance when strong ca\-tegorical-continuous
  interactions are
  present, as their underlying search distributions assume independence between
  categorical and continuous variables. To address this limitation, we propose a
  bilevel optimization framework that explicitly captures such interactions by
  optimizing over categorical variables in an outer loop, and over continuous
  variables conditioned on each categorical configuration in an inner loop. We
  formulate each level of the bilevel problem as a stochastic relaxation under
  information-geometric optimization. To mitigate the high computational cost
  inherent to bilevel optimization, we introduce a warm-starting strategy that
  accelerates the lower-level search by selecting the best among multiple cached
  configurations and updating the cache after each iteration.
  Experimental results on binary-continuous domain demonstrate that
  the proposed method outperforms existing state-of-the-art
  approaches in interaction-handling capability while also being more
  computationally efficient across benchmarks encompassing both
  previously reported and newly proposed types of interaction.

\end{abstract}
\begingroup
\makeatletter
\renewcommand\thefootnote{}
\renewcommand\@makefntext[1]{\noindent #1}
\footnotetext{The code for this paper is available at
  \url{https://www.github.com/akimotolab/igbd}; the supplementary
material is published at \url{https://doi.org/10.5281/zenodo.20623937}.}
\makeatother
\endgroup
\setcounter{footnote}{0}  %
\section{Introduction}
Mixed-variable optimization involves minimizing or maximizing an
objective function over a heterogeneous domain comprising continuous,
integer, and categorical variables. This study focuses on the
mixed-categorical case, which deals with the combination of
categorical and continuous variables:
\begin{equation}
  \min_{\cc,\xx} f(\cc,\xx),
\end{equation}
where $f: \mcc \times \mcx \to \R$, $\mcx = \R^{d_x}$, and $\mcc =
\mcc_1 \times \cdots \times \mcc_{d_c}$, with $\mcc_i$ denoting the
set of $K_i$-dimensional one-hot vectors:
\begin{equation}
  \mcc_i = \left\{\cc_i \in \{0,1\}^{K_i}\;\middle|\;\sum_{k=1}^{K_i}
  [\cc_i]_k = 1\right\}.
\end{equation}
Mixed-categorical optimization arises in many real-world
applications, including supply chain
logistics~\cite{ghodsypour2001total,steadieseifi2014multimodal,you2008mixed},
structural
optimization~\cite{barjhoux2020bi,deb2001design,kaveh2009particle},
and automated machine
learning~\cite{akimoto2019adaptive,bergstra2011algorithms,liu2018darts,ru2020bayesian}.
These problems are computationally challenging, as they fall under
the category of mixed-integer nonlinear programming (MINLP), which
contains mixed-integer linear programming (MILP) as a subset and is
therefore NP-hard in general~\cite{karp1972reducibility,koppe2011complexity}.

This study further specializes to mixed-categorical black-box
optimization (MC-BBO), where no \textit{a priori} information about
the objective is available. Existing MC-BBO approaches fall into two
broad categories: Bayesian
optimization~\cite{daxberger2021mixed,bergstra2011algorithms,ru2020bayesian,wan2021think}
and evolution
strategies~\cite{akimoto2025challenges,hamano2024catcma,li2013mixed}.
Bayesian methods are generally more sample-efficient, but scalability
limitations of their underlying surrogate models hinder application
to higher-dimensional problems, where evolution strategies tend to be
preferred. Commonly used Bayesian methods include the tree-structured
Parzen estimator (TPE)~\cite{bergstra2011algorithms} and
\textsc{Casmopolitan}~\cite{wan2021think}. Among evolution
strategies, CatCMA~\cite{hamano2024catcma} and
ICatCMA~\cite{akimoto2025challenges} represent the current state of
the art, extending the capabilities of
CMA-ES~\cite{hansen2001completely,hansen2003reducing,hansen2010comparing,akimoto2020diagonal}
to mixed-variable spaces. CatCMA models the search distribution as a
product of a multivariate Gaussian over $\xx$ and a categorical
distribution over $\cc$. The parameters of this joint distribution
are updated along their natural gradients~\cite{amari1998natural},
estimated via Monte Carlo sampling under the information-geometric
optimization (IGO) framework~\cite{ollivier2017information}, with
step-size adaptation borrowed from the adaptive stochastic natural
gradient (ASNG) method~\cite{akimoto2019adaptive} and CMA-ES.

Due to the superior scalability of evolutionary approaches, CatCMA outperforms
TPE and \textsc{Casmopolitan} on preliminary benchmarks. However, its product
distribution assumes independence between categorical and continuous variables,
preventing it from learning their interactions. To address this, ICatCMA
includes a parametrized affine map known as a hyper-representation
from the categorical to the continuous
space within the joint distribution, optimizing its parameters
directly via CatCMA. While this improves performance over a vanilla
CatCMA baseline when
categorical-continuous interactions are present, the underlying product
distribution formulation of CatCMA fundamentally limits the ability of such
approaches to capture more complex interactions. Effectively modeling such
interactions thus remains an important open problem.

A natural strategy for handling discrete-continuous interaction is to impose a
bilevel structure on the problem, since mixed-categorical optimization can be
viewed as a special case of bilevel optimization; moreover, it naturally
captures dependencies that arise in real-world problems, where different choices
of categorical variables correspond to different continuous
optimization problems, affecting both the active continuous variables and their
optimal values~\cite{audet1997links,daxberger2021mixed}. In such approaches, an
inner loop optimizes the continuous variable for a fixed categorical variable,
and an outer loop optimizes the categorical variable:\footnote{The optimal value
  of $\cc$ does not depend on the choice of $\xs$, even if multiple lower-level
optima exist.}
\begin{equation}\label{eqn:mcasbilevel}
  \begin{aligned}
    \min_{\cc} \quad &f(\cc, \xs) \\
    \mathrm{s.t.}  \quad &\xs = \argmin_{\xx} f(\cc,\xx).
  \end{aligned}
\end{equation}
Bilevel optimization is itself costly, as even linear bilevel problems in
continuous domains are NP-hard, and applying evolution strategies to bilevel
optimization is an active research
area~\cite{chen2024evolutionary,chen2025motea,he2018evolutionary,huang2023bilevel,ong2026accelerating}.
A key drawback of evolutionary approaches is their nested loop structure, which
requires solving the lower-level problem to convergence for every upper-level
candidate at each iteration. Due to the the high computational cost this
entails, no prior work, to our knowledge, has applied such bilevel approaches to
MC-BBO within the framework of ES.

In the continuous setting, \cite{ong2026accelerating} mitigates this cost by
augmenting nested CMA-ES optimizers with heuristics to accelerate bilevel
optimization. One such heuristic is a warm-starting strategy that
maintains a cache
of multiple lower-level configurations; at each iteration, the best
cached configuration
is selected for each upper-level candidate and subsequently updated after the
lower-level search. This yields significant gains in both efficiency and
robustness on problems with strong interactions and multimodality.
Although \cite{ong2026accelerating} considers only the continuous domain at both
levels, its warm-starting strategy generalizes readily to any rank-based
upper-level solver and any stateful lower-level solver; we therefore expect
it to be effective for handling interactions in the bilevel formulation of
MC-BBO.

\paragraph{\textup{\textbf{Contributions}}} The contributions of this
paper are as follows:

\begin{enumerate}
  \item We propose mixed-categorical optimization via \proposedmethod
    (\prme), which formulates each level of~\Cref{eqn:mcasbilevel} as
    a stochastic relaxation under IGO and incorporates warm-starting
    via a cache of multiple configurations,
    following~\cite{ong2026accelerating}.

  \item We introduce a novel type of categorical-continuous
    interaction in which categorical variables govern the
    interactions between continuous variables.

  \item We benchmark \prme against CatCMA and ICatCMA on a range of
    both existing and newly proposed benchmark problems in the
    binary-continuous domain,\footnote{Due to space limitations, this
      paper considers only the binary-continuous domain, although the
      proposed approach extends naturally to general mixed-categorical
      optimization in principle. Investigating the computational cost
      and practical feasibility of scaling to a larger number of
    categories is an important direction for future work.}
    demonstrating improvements in both computational cost and
    interac\-tion-handling effectiveness.
\end{enumerate}

\section{Proposed Method}\label{s:prme}

\subsection{Bilevel Stochastic Relaxation under IGO}
The IGO framework~\cite{ollivier2017information} provides a recipe
for constructing rank-based evolutionary algorithms for black-box
optimization via stochastic relaxation. Given an objective $f:\mcx
\to \R$, IGO minimizes its expectation
$J(\ttheta)\triangleq\E_{\xx\sim p_{\ttheta}}[f(\xx)]$ under a
parametrized distribution $p_{\ttheta}$ by iteratively updating
$\ttheta$ along a Monte-Carlo estimate of the natural gradient,
concentrating the distribution around the optimum $\xs$.

We formulate each level of the bilevel problem
in~\Cref{eqn:mcasbilevel} as a stochastic relaxation under IGO, using
a categorical distribution for the discrete upper level and a
multivariate Gaussian for the continuous lower level. We further
incorporate the adaptation mechanisms of the ASNG method at the upper
level and CMA-ES at the lower level.

\subsubsection{Upper Level}
Let $\qq \triangleq [\qq_1, \ldots, \qq_{d_c}]\T$, where $\qq_i$ is
the probability vector governing the distribution of $[\cc]_i$. The
upper-level stochastic relaxation is
\begin{equation}\label{eqn:ulsr}
  \min_{\qq} J_{f^\star}(\qq),\ \textnormal{where }J_{f^\star}(\qq)
  \triangleq \E_{\cc \sim \Cat(\qq)}[f^\star(\cc)] =
  \sum_{\cc\in\mcc} f(\cc) p(\cc;\qq),
\end{equation}
with upper-level objective $f^\star(\cc) \triangleq f(\cc,\xs(\cc))$.
The natural gradient with respect to $\qq$ is obtained by
premultiplying the vanilla gradient by the inverse Fisher information matrix:
\begin{align}
  \tilde \nabla_{\qq} J_{f^\star}(\qq) = \FF^{-1}(\qq) \nabla
  J_{f^\star}(\qq) = \FF^{-1}(\qq) \E_{\cc\sim\Cat(\qq)}[f^\star
  (\cc) \nabla_{\qq} \log p(\cc;\qq)],
\end{align}
where $\FF(\qq) \triangleq \E_{\cc\sim\Cat(\qq)}[\nabla_{\qq} \log
p(\cc; \qq)\nabla_{\qq} \log p(\cc; \qq)\T]$ is the Fisher
information matrix. Its Monte Carlo estimate $\gq$ using $\lambda_c$ samples is
\begin{align}\label{eqn:rawgq}
  \gq = \frac{1}{\lambda_c} \FF^{-1}(\qq) \sum_{i=1}^{\lambda_c}
  f^\star (\cc_i) \nabla_{\qq} \log p(\cc_i;\qq)  =
  \frac{1}{\lambda_c} \sum_{i=1}^{\lambda_c} f^\star(\cc_i) (\cc_i-\qq).
\end{align}
IGO employs fitness shaping, replacing $f^\star(\cc)$ with rank-based
weights $w_i$ (larger positive weights correspond to lower objective
values) to ensure invariance to monotonic transformations, giving the update
\begin{equation}\label{eqn:asngupdate}
  \qq\gets\qq+\eta\gq,\ \mathrm{where}\ \gq =\frac{1}{\lambda_c}
  \sum_{i=1}^{\lambda_c} w_i (\cc_{i:\lambda_c}-\qq).
\end{equation}
We adopt ASNG for the upper-level optimization, which
casts~\Cref{eqn:asngupdate} as a trust region problem under the
Fisher information metric, setting $\eta =
\delta/\|\gq\|_{\FF(\qq)}$, where $\|\cdot\|_{\FF(\qq)}$ is the
Fisher norm and $\delta$ is the trust radius. The trust radius is
adapted analogously to cumulative step-size adaptation
(CSA)~\cite{hansen2003reducing}:
\begin{equation}\label{eqn:trustregionadaptation}
  \begin{aligned}
    \beta &\gets \frac{\delta}{\sqrt {\sum_{i=1}^{d_c} (K_i-1)}};\
    \sS \gets (1 - \beta) \sS +
    \sqrt{\beta(2-\beta)}\frac{\FF^{1/2}(\qq) \gq}{\|\gq\|_{\FF(\qq)}};\\
    \gamma &\gets (1 - \beta)^2  \gamma + \beta(2-\beta);\ \delta
    \gets \delta \exp\left(\beta\left(\frac{\|\sS\|^2}{\alpha} -
    \gamma\right)\right),
  \end{aligned}
\end{equation}
where $\beta$ is the cumulation rate, $\sS$ is a quantity analogous
to the evolution path in CSA, $\gamma$ is the expected value of
$\|\sS\|^2$ under noise, and $\alpha$ controls the target signal-to-noise ratio.
To prevent the distribution from becoming degenerate, $\qq$ is then
clipped such that
\begin{equation}\label{eqn:clipq}
  \frac{1}{d_c(K_i-1)} \le [\qq_i]_k \le 1 -
  \frac{1}{d_c(K_i-1)},\quad\forall i\in\{1,\ldots,d_c\},\ \forall
  k\in\{1,\ldots,K_i-1\}.
\end{equation}
We also adopt the ASNG rank-based weighting scheme:
\begin{equation}
  w_i =
  \begin{cases}
    +1,&\mathrm{if}\ 1\le i\le \lceil \lambda/4 \rceil,\\
    -1,&\mathrm{if}\ \lambda -\lceil \lambda/4 \rceil +1 \le i \le \lambda,\\
    0,&\mathrm{otherwise}.
  \end{cases}
\end{equation}

\subsubsection{Lower Level}
We minimize $J_{f_{\cc}}(\oomega)$, the expectation of $f_{\cc}(\xx)
\triangleq f(\cc,\xx)$ under a multivariate Gaussian parametrized by
$\oomega \triangleq [\mm\T, \vecop(\SSigma)\T]\T$:
\begin{equation}\label{eqn:llsr}
  \min_{\oomega} J_{f_{\cc}}(\oomega),\ \textnormal{where
  }J_{f_{\cc}}(\oomega) \triangleq \E_{\xx \sim
  \mcn(\mm,\SSigma)}[f_{\cc}(\xx)] = \int_{\R^{d_x}} f_{\cc}(\xx)
  p(\xx;\mm,\SSigma)\,\rmd\xx.
\end{equation}
In the multivariate Gaussian case, the IGO parameter update analogous
to~\Cref{eqn:asngupdate} based on Monte Carlo estimation of the
natural gradient of $J_{f_{\cc}}(\oomega)$ using $\lambda_x$ samples
is known to correspond to the so-called rank-$\mu$ update of
CMA-ES~\cite{akimoto2010bidirectional,ollivier2017information}:
\begin{align}
  \mm &\gets \mm + \eta \sum_{i=1}^{\lambda_x} w_i(\xx_{i:\lambda_x}-\mm);\\
  \SSigma &\gets \SSigma + \eta\sum_{i=1}^{\lambda_x}
  w_i[(\xx_{i:\lambda_x}-\mm)(\xx_{i:\lambda_x}-\mm)\T - \SSigma].
\end{align}
In practice, the CMA-ES algorithm commonly used today incorporates additional
heuristically motivated update rules beyond the rank-$\mu$ update, most notably
the rank-one update of the covariance matrix and cumulative step-size
adaptation~\cite{hansen2001completely}. Within our proposed method, we
adopt a recent variant known as dd-CMA-ES~\cite{akimoto2020diagonal}
for the lower-level optimization, which
further incorporates the active update~\cite{jastrebski2006improving} and
adaptive diagonal decoding.

\subsection{Lower-Level Optimization with Warm Starting}
\begin{algorithm}[t]
  \caption{Upper-level categorical optimization with ASNG}\label{alg:upperlevel}
  \begin{algorithmic}[1]
    \Require Initial $\qq$, hyperparameters $\lambda_c = 8$, $\alpha=3/2$
    \State $\delta \gets 1$, $\gamma \gets 0$, $\sS \gets \Zero$
    \While{not converged}
    \State $\{\cc_1,\ldots, \cc_{\lambda_c}\} \sim \Cat(\qq)$
    \State Obtain rankings
    $\{\cc_{1:\lambda_c},\ldots,\cc_{\lambda_c:\lambda_c}\}$ using
    \Cref{alg:WRA}
    \State $\gq \gets \frac{1}{\lambda_c} \sum_{i=1}^{\lambda_c} w_i
    (\cc_{i:\lambda_c}-\qq)$, $\eta \gets \delta/\|\gq\|_{\FF(\qq)}$,
    $\qq \gets \qq + \eta \gq$
    \State Adapt trust region parameters $\delta$, $\gamma$, $\sS$
    according to~\Cref{eqn:trustregionadaptation}
    \State Clip $\qq$ according to~\Cref{eqn:clipq}
    \EndWhile
  \end{algorithmic}
\end{algorithm}
\begin{algorithm}[t]
  \caption{Lower-level optimization with warm starting}\label{alg:WRA}
  \begin{algorithmic}[1]
    \Require $\cc_1, \dots, \cc_{\lambda_c}$
    \Require $\{(\xx_k, \oomega_k, p_k)\}_{k=1}^{\ny}$
    \Require $\pthr$, $\pp$, $\pn$

    \State\label{line:WRA-initstart}\texttt{// Configuration selection}
    \For{$i \in \{ 1, \dots, \lambda_c\}$}
    \State Evaluate $f(\cc_{i}, \xx_k)$ for all $k \in \{1,\dots,\ny\}$
    \State $k^\star_i \gets \argmin_k f(\cc_i,\xx_k)$
    \State $\xh_i\gets\xx_{k^\star_{i}}$,
    $\oomegat_i\gets\oomega_{k^\star_{i}}$, $\tilde f_i\gets f(\cc_i,
    \xx_{k^\star_{i}})$
    \EndFor\label{line:WRA-initend}

    \State\label{line:WRA-worststart}\texttt{// Lower-level optimization}
    \For{$i \in \{1, \dots, \lambda_c\}$}
    \State $\tilde f_{i}, \xh_i, \oomegat_i, \gets
    \textsc{LowerLevel}(\tilde f_{i}, \xh_i, \oomegat_i)
    $\texttt{\ \ // \Cref{alg:CMAESinWRA}}
    \EndFor

    \State\texttt{// Post-processing}\label{line:WRA-poststart}
    \State $S^\star \gets \{{k^\star_{i}}\}_{i=1}^{\lambda_c}$
    \For{$\tilde{k} \in S^\star$}
    \State $\ell \gets \argmin_{i=1,\dots,\lambda_c} \{\tilde f_i
    \mid k^\star_i = \tilde{k}\}$
    \State $\xx_{\tilde{k}} \gets \xh_\ell$, $\oomega_{\tilde{k}}
    \gets \oomegat_\ell$, $p_{\tilde{k}} \gets \min(p_{\tilde{k}} +
    \bar{p}_{+}, 1)$
    \EndFor
    \State $p_k \gets p_k - \bar{p}_{-} \cdot \ind{k \notin S^\star}$
    for $k\in\{1,\ldots,\ny\}$
    \For{$k \in \{1,\dots,\ny\}$}
    \State Refresh $(\xx_k, \oomega_k, p_k)$ \textbf{if} $p_k <
    p_{\mathrm{threshold}}$
    \EndFor
    \label{line:WRA-postend}
    \State \Return $\{\tilde f_i\}_{i=1}^{\lambda_c}$ and $\{(\xx_k,
    \oomega_k, p_k)\}_{k=1}^{\ny}$ for the next call
  \end{algorithmic}
\end{algorithm}
\begin{algorithm}[t]
  \caption{Lower-level optimization with CMA-ES}\label{alg:CMAESinWRA}
  \begin{algorithmic}[1]%
    \Require $\cc, \xh, f_x, \oomegat = (\mt, \SSigmat)$  %
    \Require $\Tmin$, $\Vxmin >0$, $\lambda_x = \lfloor 4 + 3\log d_x  \rfloor$

    \State $\SSigmat_{\mathrm{init}} \gets \SSigmat$, $t' \gets 0$  %
    \While{$h = \textsc{False}$}

    \State Sample $ \{\xhp\}_{k=1}^{\lambda_x} \sim \mathcal{N}(\mt, \SSigmat)$
    \State Evaluate $f_k \gets f(\cc, \xhp_{k})$ for all
    $k\in\{1,\ldots,\lambda_x\}$

    \State Select the best index
    $\tilde{k}^{\mathrm{min}}\gets\argmin_{k\in\{1,\dots,\lambda_x\}} f_k$
    \If {$f(\cc, \xhp_{\tilde{k}^{\mathrm{min}}}) \le f_x$}
    \State $f_x \gets \min_{k\in\{1,\dots,\lambda_x\}} f_k$, $\xh
    \gets \xhp_{\tilde{k}^{\mathrm{min}}}$
    \EndIf

    \State Perform dd-CMA-ES update using $\{\xhp_{k}, f_k\}_{k=1}^{\lambda_x}$
    \If {$\max_\ell\sqrt{[\SSigmat]_{\ell,\ell}} < \Vxmin$ \text{and}
    $t' \geq \Tmin$}
    \State\label{l:sig_correct_1}$\DD \gets \diag\bigg(\max\bigg(1,
      \frac{\Vxmin}{\sqrt{[\SSigmat]_{1,1}}}\bigg), \dots, \max\bigg(1,
    \frac{\Vxmin}{\sqrt{[\SSigmat]_{d_x,d_x}}}\bigg)\bigg)$

    \State\label{l:sig_correct_2}$\SSigmat \gets \DD \SSigmat \DD$
    and $h \gets \textsc{True}$
    \EndIf
    \State Set $\SSigmat \gets \SSigmat_{\mathrm{init}}$ and $h \gets
    \textsc{True}$ \textbf{if} $\kappa(\SSigmat) > \Cxmax$
    \State $t' \gets t'+1$
    \EndWhile
    \State \Return $\xh$, $f_x$, $\oomegat = (\mt, \SSigmat)$  %
  \end{algorithmic}
\end{algorithm}

Combining ASNG at the upper level with CMA-ES at the lower level following the
hierarchical structure of~\Cref{eqn:mcasbilevel} yields a bilevel algorithm that
can in principle solve MC-BBO problems. However, running CMA-ES until
convergence for each ASNG iteration becomes prohibitively expensive. To remedy
this, we augment the lower-level CMA-ES search with the warm-starting strategy,
which passes the approximated optimal continuous variables for each
categorical variable to the upper-level ASNG solver
(\Cref{alg:upperlevel}).

The details of the warm-starting procedure are shown in \Cref{alg:WRA}. Given a
set of $\lambda_c$ upper-level candidate solutions $\{\cc_i\}_{i=1}^{\lambda_c}$
and a cache of $N_\omega$ lower-level solver configurations $\{(\xx_k,
\oomega_k, p_k)\}_{k=1}^{N_\omega}$, where $p_k \in [0,1]$ is a score measuring
the quality of a configuration, the procedure operates in three phases. First, a
warm-starting phase selects, for each candidate $\cc_i$, the best cache
configuration $k_i^\star$, using it to initialize the corresponding lower-level
solver. Then, in the lower-level optimization phase, a lower-level CMA-ES solver
(\Cref{alg:CMAESinWRA}) is run for each candidate, updating the estimate
$\tilde{f}(\cc_i)$ of $f^\star(\cc_i)$ until a termination condition based on
either the condition number of the covariance matrix or the maximum
coordinate-wise standard deviation is met.  %

Finally, a post-processing phase updates the cache with the newly
learned solver configurations.
Each cache entry selected by at least one lower-level solver is
overwritten with the configuration
of the solver whose corresponding upper-level candidate achieved the
best estimated
$\tilde{f}(\cc_i)$. To promote exploration and prevent stale
configurations from persisting,
the score $p_k$ is incremented by $\pp$ when entry $k$ is selected
and decremented by $\pn$
otherwise; entries whose score falls below $\pthr$ are refreshed by
re-initializing to a new random CMA-ES distribution.
The procedure returns the estimate $\xh$ of $\xs$ for each candidate
of the upper-level population and an updated cache.  %

\section{Benchmark Problems}\label{s:benchmarkproblems}

In this section, we introduce the benchmark problems used to evaluate \prme
against the CatCMA and ICatCMA baselines. We consider the three interaction
types proposed in~\cite{akimoto2025challenges}, and additionally propose a
fourth type in which interactions within the continuous domain depend strongly
on $\cc$, a challenge absent from the existing three types, designed to mimic
real-world scenarios where different values of $\cc$ give rise to substantially
different optimization problems. Following~\cite{akimoto2025challenges}, we
restrict all benchmark problems to binary discrete variables, i.e. $K_i=2$ for
$i\in\{1,\ldots,d_c\}$. As such, for the remainder of this paper, we adopt a
slight abuse of notation: $\cc \in \{0,1\}^{d_c}$ denotes a bitstring of length
$d_c$ rather than a list of $d_c$ one-hot vectors of length $2$, and
$\qq\in[0,1]^{d_c}$ denotes a single probability vector of a factored Bernoulli
distribution on $\{0,1\}^{d_c}$, rather than a list of probability vectors as in
\Cref{s:prme}.

\subsection{Type-I Interaction}

Type-I interaction characterizes scenarios where $\cc$ determines the
active components of $\xx$.  Assuming $d_c=d_x$, the benchmark problem is
\begin{equation}\label{eqn:fI}
  \fI(\cc, \xx) =  \|\xx \odot \cc - \bs\|^2,
\end{equation}
where $\odot$ denotes the Hadamard product and $\bs \in \R^{d_x}$ is
a fixed vector. For any given $\cc$, the optimal continuous vector is
$\xs(\cc)=\bs$,\footnote{The solution is not unique if $\cc$ contains
zeros.} and the Hessian with respect to $\xx$ is $\HH=2\diag(\cc)$,
which is rank-deficient when $\cc$ contains zeros.

The optimal binary component $[\cs(\xx)]_i$ flips depending on the
deviation of $[\xx]_i$ from $[\bs]_i$ relative to the magnitude of $[\bs]_i$:
\begin{equation}
  [\cs(\xx)]_i = \ind{ |[\xx-\bs]_i| \le |[\bs]_i|}.
\end{equation}
The global optimum is $(\cs, \xs) = (\One, \bs)$,\footnote{The
solution is not unique if $\bs$ contains zeros.} and solving the
problem requires flipping all bits of $\cc$ to 1.

\subsection{Type-II Interaction}
Type-II interaction characterizes cases where $\cc$ determines the
optimal location of $\xx$ via a mapping $\pphis : \mcc \to \mcx$. The
benchmark problem is
\begin{equation}\label{eqn:fII}
  \fII(\cc, \xx) = \om(\cc) + \|\xx-\pphis(\cc)\|^2,
\end{equation}
where $\om(\cc) \triangleq \sum_{i=1}^{d_c}(1-[\cc]_i)$. For a fixed
$\cc$, the problem reduces to a spherical function in $\xx$ with
optimum $\xs(\cc)=\pphis(\cc)$ and isotropic Hessian $\HH=2\II$.

For simplicity, $\pphis$ is defined as the affine map $\pphis(\cc) =
a\Vs\cc+\bs$,
where $\Vs \in \R^{d_x \times d_c}$, $\bs \in \R^{d_x}$, and $a \ge
0$. The parameter $a$ controls the strength of the binary-continuous
interaction. As $a \to 0$, the problem approaches separability: $\fII
= \om(\cc) + \|\xx-\bs\|^2$, giving $\cs(\xx)=\One$. For
non-negligibly large $a$, both terms become significant and
determining $\cs(\xx)$ becomes a nontrivial binary quadratic
programming (BQP) problem. %
The global optimum is $(\cs,\xs)=(\One,a\Vs\One + \bs)$.

\subsection{Type-III Interaction}
Type-III interaction combines the difficulties of types I and II:
$\cc$ simultaneously masks the active components of $\xx$ and
determines the optimum $\xs(\cc)$. Assuming $d_c=d_x$, the benchmark problem is

\begin{equation}\label{eqn:fIII}
  \fIII(\cc, \xx) = \|\xx \odot \cc - \pphis(\cc)\|^2,
\end{equation}
using the same affine map $\pphis(\cc)$ as in $\fII$. For a fixed
$\cc$, the optimal continuous vector\footnote{The solution is not
unique if $\pphis(\cc)$ contains zeros.} is $\xs(\cc)=\pphis(\cc)$
and the Hessian is $\HH=2\diag(\cc)$. As with $\fII$, $a$ controls
the interaction strength: as $a\to0$,  the problem reduces to $\fI$,
while for sufficiently large $a$ determining $\cs(\xx)$ becomes a BQP
problem.  %
The global optimum is $(\cs,\xs)=(\One,a\Vs\One + \bs)$.

\subsection{Type-IV Interaction}
For all three interaction types, a single isotropic covariance matrix
still suffices for CMA-ES to efficiently optimize $\xx$ regardless of
$\cc$: the Hessian is $2\II$ for $\fII$, and $2\diag(\cc)$ for $\fI$
and $\fIII$.
In many real-world applications, however, the optimal distribution of
$\xx$ can vary significantly with $\cc$, a dependency intractable
with product-distribution-based approaches.

To address this, we introduce type-IV variants of $\fI$, $\fII$, and
$\fIII$, which we denote as $\fIpIV$, $\fIIpIV$, $\fIIIpIV$,
respectively, by replacing the Euclidean norm
in~\Cref{eqn:fI,eqn:fII,eqn:fIII} with a Mahalanobis norm
$\|\xx\|^2_{\Aa(\cc)}\triangleq\xx^\top \Aa(\cc) \xx$ dependent on
$\cc$, with the map $\Aa : \mathcal{C} \to \Spp^{d_x}$. The resulting
Hessian then depends on $\cc$ as  $\HH = 2\Aa(\cc)$ for $\fIIpIV$ and
$\HH = 2\diag(\cc)\Aa(\cc)\diag(\cc)$ for $\fIpIV$ and $\fIIIpIV$.
Moreover, this modification leaves $\xs(\cc)$ and the global optimum
$(\cs,\xs)$ unchanged, though the BQP parameters determining
$\cs(\xx)$ differ from the original $\fII$ and $\fIII$. The
combination of types II and III with type IV results in problems
where both $\xs$ and the interactions within the continuous domain
depend on $\cc$.

We set $\Aa(\cc)=\QQ(\cc) \LLambda \QQ(\cc)^\top$, where
$\LLambda\in\mathbb{R}^{d_x\times d_x}$ is a fixed diagonal matrix of
positive eigenvalues and $\QQ(\cc)\in\mathbb{R}^{d_x \times d_x}$ is
the orthogonal matrix
\begin{equation}
  \QQ(\cc) = \exp \left( \MM_0 + \sum_{i=1}^{d_c} [\cc]_i \MM_i \right),
\end{equation}
with fixed skew-symmetric matrices $\MM_0,\MM_1,\ldots,\MM_{d_c} \in
\mathbb{R}^{d_x \times d_x}$ and $\exp(\cdot)$ the matrix
exponential. The condition number $\kappa(\Aa)=\kappa(\LLambda)$
controls the strength of the type-IV interaction: as
$\kappa(\Aa)\to1$, $\LLambda\propto\II$ and $\Aa(\cc)\propto\II$,
causing the type-IV interaction to vanish.
\section{Experiments}
To evaluate the computational efficiency, interaction-handling
ability, and scalability of \prme, we benchmark it against CatCMA and
ICatCMA on problems with the interaction types described in
\Cref{s:benchmarkproblems} with varying dimensionalities. For each
method and problem, we perform 20 independent runs\footnote{For only
  ICatCMA on the $d_c=d_x=10$ case, we perform 10 runs due to its high
  computational cost at higher dimensionality; 20 runs are performed in
all other cases.} and report the success rate, defined as the
fraction of runs achieving an objective function value $\num{1e-6}$
or lower from the true optimum, and the total number of function
evaluations (FEs). All methods share the same termination conditions,
and we follow the baseline experimental settings and hyperparameters
recommended in the original CatCMA and ICatCMA papers.

\subsection{Benchmark Problems}
The test suite used to benchmark our proposed method against the baselines
comprises all binary-continuous problems described in
\Cref{s:benchmarkproblems}, namely $\fII$, $\fIII$, and their type-IV variants
$\fIIpIV$ and $\fIIIpIV$.\footnote{
  $\fI$ and $\fIpIV$ are included as cases of $\fIII$ and $\fIIIpIV$ when $a=0$.
  $\fIIpIV$ when $a=0$ corresponds to pure type-IV interaction; its
  objective function is given by
  $\om(\cc)+(\xx-\bs)\T\Aa(\cc)(\xx-\bs)$; the second term
  corresponds to a rotated ellipsoid centered at $\bs$.
}
The constants used for each of
these problems are set as follows:

\paragraph{\textup{\textbf{Optimal Mapping}}}
Following~\cite{akimoto2025challenges}, for the optimal mapping $\pphis(\cc)$,
$\Vs$ is generated as $[\Vsh]_{i,j} \sim \mcn(0, 1)$ and then subsequently
normalized as $\Vs = \Vsh/\|\Vsh\|_F$. Likewise, $\bs$ is initialized as
$[\bsh]_i \sim \mcn(0,1)$ and $\bs = \bsh/\|\bsh\|_2$. We vary the interaction
strength as $a \in \{0,1,2,4,8,16\}$.

\paragraph{\textup{\textbf{Type-IV Interaction}}}
To control the strength of type-IV interaction, we consider both a
well-conditioned case, where the $d_x$ eigenvalues of $\Aa(\cc)$ are
logarithmically equispaced over $[10^0,10^2]$, yielding a condition
number $\kappa(\Aa)=10^2$, and an ill-conditioned case, where the
interval is $[10^0, 10^6]$ and $\kappa(\Aa)=10^6$.
$\{\MM_i\}_{i=0}^{d_c}$ is generated as $[\Mh_i]_{j,k}\sim \mcn(0,1)$
and $\MM_i = 20(\Mh_i - \Mh{}_i^\top)$.

\subsection{Initial Distribution Parameters and Hyperparameters}

\paragraph{\textup{\textbf{CatCMA and ICatCMA}}} The search
distribution parameters are
initialized as $\qq=[1/2,\ldots,1/2]\T$, $\mm=\Zero$, and $\SSigma=\II$. All
hyperparameters, including those controlling termination criteria based on
numerical properties of the search distribution, follow the
recommended values from their respective
studies~\cite{hamano2024catcma,akimoto2025challenges}.

\paragraph{\textup{\textbf{\prme}}}  For the upper level ASNG solver,
the search distribution
is initialized as $\qq=[1/2,\ldots,1/2]\T$ with $\delta=1$,
$\gamma=0$, and $\sS=\Zero$. The initial
lower-level warm-starting cache configurations $\{(\xx_k, \oomega_k,
p_k)\}_{k=1}^{\ny}$ are $\xx_k\sim \mcu[0,1]^{d_x}$, $\oomega_k =(\mm_k,
\SSigma_k)=(\Zero,\II)$, and $p_k=1$. The hyperparameters for ASNG
(\Cref{alg:upperlevel}) are $\lambda_c=8$ and $\alpha=3/2$, the same
as those used within the ASNG categorical search of CatCMA and ICatCMA. For warm
starting (\Cref{alg:WRA}), we use $\ny=3\times\lambda_c$,
$\pthr=0.1$, $\pp=0.4$, and
$\pn=0.05$; for CMA-ES (\Cref{alg:CMAESinWRA}), $\lambda_x=\lfloor 4+ 3 \log
d_x\rfloor$,  $\Tmin=10$, $\Vxmin=10^{-4}$, and $\Cxmax=10^{7}$.

\subsection{Termination and Restart Strategy}

In addition to termination criteria based on the numerical properties of the
search distribution, we adopt budget-based termination.
Specifically, the algorithm is terminated if no improvement in the objective
function greater than $\num{1e-6}$ is observed after 50 generations of the
upper-level ASNG (\Cref{alg:upperlevel}) for \prme, or after 50 generations for
the CatCMA and ICatCMA baselines. For \prme, the lower level
(\Cref{alg:CMAESinWRA}) is also terminated if no improvement greater than
$\num{1e-6}$ is observed after 20 generations. Upon termination, a simple
restart strategy resets all distribution parameters; in the case of
\prme, the parameters of both levels of the optimization are reset.
Restarts continue until any of
the following conditions is met: the objective function reaches within
$\num{1e-6}$ of the true optimum or the total
number of FEs across restarts exceeds $\num{2e4}\times d_c \times d_x$. The best
objective function value across all restarts is reported.

\subsection{Results}\label{ss:results}

\begin{figure}[t]
  \centering
  \includegraphics[width=\linewidth]{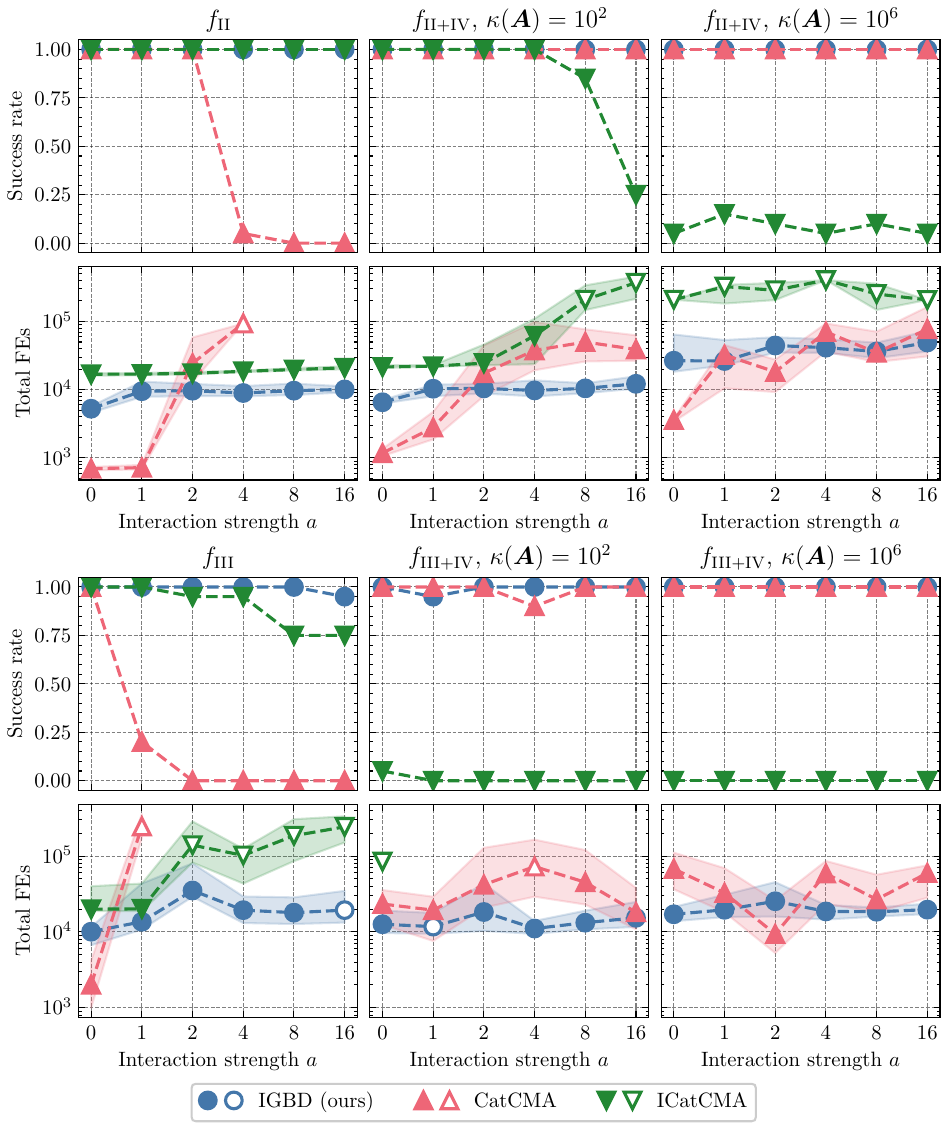}
  \caption{
    Comparison of the success rate and total function evaluations (FEs) for
    CatCMA, ICatCMA, and the proposed method on $\fII$,
    $\fIII$, and their corresponding well-conditioned and ill-conditioned
    type-IV variants with $d_c=d_x=5$ as a function of $a$. In the FE plots,
    markers and bands denote the median and interquartile range (IQR)
    of FEs, respectively, calculated only over successful runs.
    Filled markers (\CIRCLE) indicate a
    100\% success rate; hollow markers (\Circle) indicate a lower success
    rate. If a method was unsuccessful for all runs, its median FE
    count is omitted.
  }

  \label{fig:5_5}
\end{figure}
\begin{figure}[t]
  \centering
  \includegraphics[width=\linewidth]{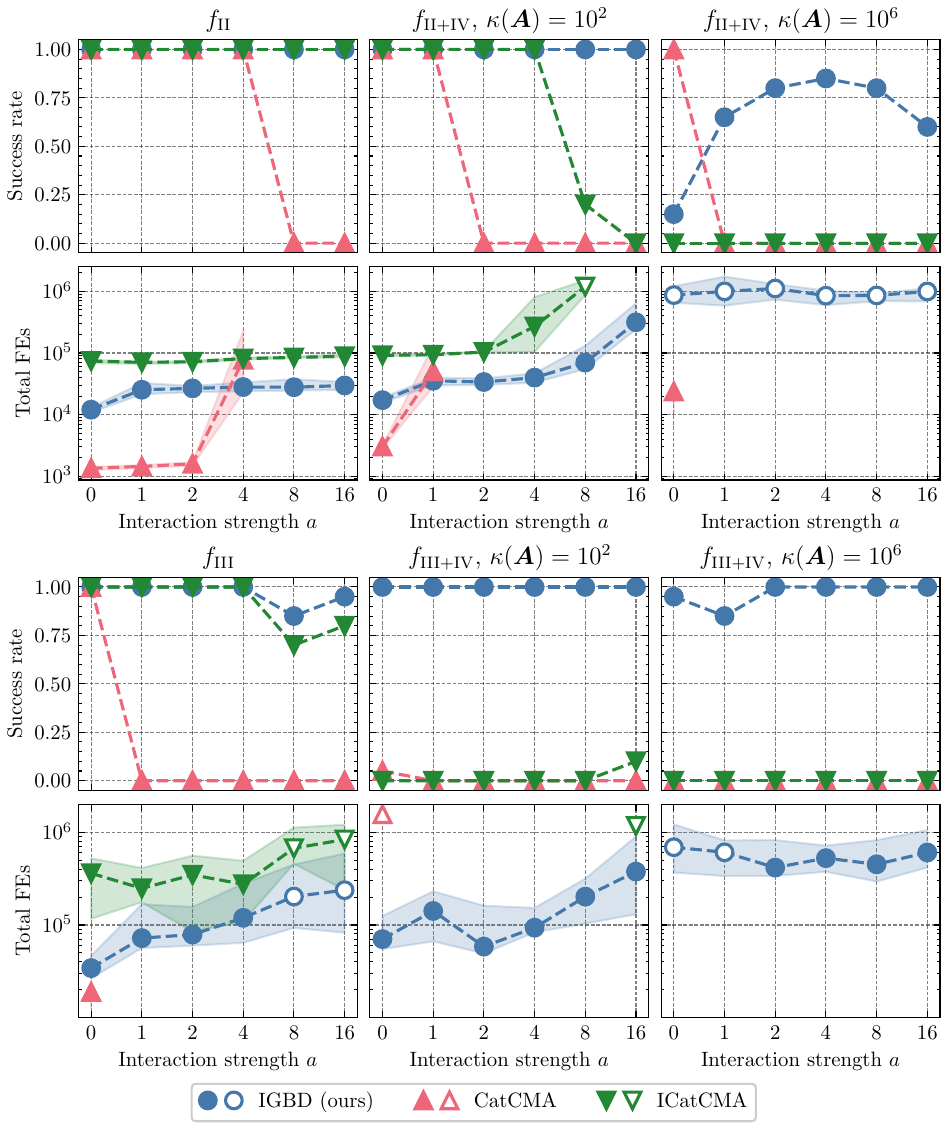}
  \caption{ Comparison of the success rate and total function evaluations (FEs)
    for CatCMA, ICatCMA, and the proposed method on $\fII$,
    $\fIII$, and their corresponding well-conditioned and ill-conditioned
    type-IV variants as a function of $a$ with $d_c=d_x=10$.
  Markers and bands in FE plots follow the conventions of \Cref{fig:5_5}.}
  \label{fig:10_10}
\end{figure}

\Cref{fig:5_5} shows the performance of \prme, CatCMA, and ICatCMA on $\fII$,
$\fIII$, and their type-IV variants when $d_c = d_x = 5$. \prme achieves a 100\%
success rate on all problems, with only two exceptions: $\fIII$ at $a = 16$ and
$\fIIIpIV$ at $a = 1$ when $\kappa(\Aa) = 10^2$, both of which still
reach 95\%. %
ICatCMA attains 100\% only on $\fII$ (for all $a$), on $\fIIpIV$ for $a \le 4$
when $\kappa(\Aa) = 10^2$, and on $\fIII$ for $a \in \{0, 1\}$. CatCMA reaches
100\% on $\fIIpIV$ and $\fIIIpIV$ across all settings, except on $\fIIIpIV$ at
$a = 4$ when $\kappa(\Aa) = 10^2$ (90\%); without the type-IV interaction,
however, it attains 100\% only for $a \le 2$ on $\fII$ and $a = 0$ on $\fIII$,
declining sharply thereafter. In terms of computational cost, ICatCMA is
consistently slower than both CatCMA and \prme; between the remaining
two, CatCMA
sometimes outperforms \prme for small $a$, but \prme is comparable or better
otherwise, with its advantage generally growing as $a$ increases; \prme is also
more robust than both competing approaches.\footnote{A statistical
  analysis confirming the significance of these trends, as well as
  those in the $d_c=d_x=10$ case, is presented in
\Cref{apdx:statistical} of the supplementary material.}

\Cref{fig:10_10} shows the results at the higher dimensionality $d_c = d_x =
10$. As before, ICatCMA achieves 100\% on $\fII$ for all $a$; once the type-IV
interaction is introduced, however, it succeeds 100\% only for $a \le 4$ when
$\kappa(\Aa) = 10^2$ and fails entirely for all $a$ when $\kappa(\Aa) = 10^6$.
On $\fIII$, ICatCMA reaches 100\% for $a \le 4$, but drops to near 0
on $\fIIIpIV$
for all $a > 0$ under both values of $\kappa(\Aa)$. In contrast to the $d_c =
d_x = 5$ case, $\fIIpIV$ is more challenging than $\fII$ for CatCMA, which
succeeds only for $a \le 1$ on the former versus $a \le 4$ on the latter; its
success rate also drops to 0\% for all $a > 0$ on $\fIIpIV$ with $\kappa(\Aa) =
10^6$ and on both $\fIII$ and $\fIIIpIV$ regardless of $\kappa(\Aa)$. \prme
retains 100\% across all settings, except on $\fIIpIV$ with $\kappa(\Aa)=10^6$
(for all $a$), $\fIII$ at $a \in \{8, 16\}$, and $\fIIIpIV$ at $a \in \{0, 1\}$.

We now interpret these trends. The bilevel formulation of \prme explicitly
captures dependencies between $\cc$ and $\xx$, and its relative success across
all interaction types regardless of $a$ demonstrates practical robustness when
combined with cached warm-starting, even at higher dimensionality. Its failures
on $\fIIpIV$ and $\fIIIpIV$ all occur in high-interaction regimes (large $a$ or
type-IV interaction) that demand many FEs to converge; since the median FE count
is close to the budget in these cases, the failures can be attributed primarily
to an insufficient FE budget rather than to a fundamental limitation of the
method.

By contrast, while ICatCMA reliably solves $\fII$ for all $a$, this is because
the affine optimal mapping $\xs(\cc)$ exactly matches the form of its
hyper-represen\-tation. When nonlinear type-III or type-IV interactions are
present, ICatCMA's success rate decreases markedly with interaction strength. A
further factor in ICatCMA's high cost is that its affine hyper-representation
has $\mathcal{O}(d_c d_x)$ degrees of freedom, already significant at modest
dimensionalities; increasing its expressivity to handle more complex
interactions would likely raise this cost further.

Finally, the independence assumption of CatCMA prevents it from
handling strong interactions regardless of linearity, but significantly reduces
cost when this assumption holds.
Interestingly, at $d_c=d_x=5$, CatCMA achieves higher success rates
when type-IV interaction is present. While we cannot precisely
explain why $\fII$ and $\fIII$ become easier, one partial intuition
is that CatCMA's
fast convergence leads to hundreds of restarts within the FE budget
(see \Cref{apdx:statistical} for restart statistics), effectively
introducing a random-search component where many initializations raise the
chance of a favorable start; type-IV interaction may render the search space
more amenable to this approach. At $d_c=d_x=10$, however, the
larger search space renders this
strategy ineffective, with both CatCMA
and ICatCMA dropping to near-zero
success under type-IV interaction.

\section{Conclusion}
We introduced \prme, a bilevel information-geometric approach to
mixed-catego\-rical black-box optimization with a
multi-configurational warm-starting mechanism. On the mixed
binary-continuous domain, we demonstrate that our method handles both
previously reported and novel interaction types significantly better
than current state-of-the-art methods, while also improving efficiency.

Several directions remain for future work. Increasing $K_i$ beyond
binary problems is the natural next step, as is further
taxonomization and benchmarking of interaction types. Incorporating
low-effective-dimension handling into the lower-level optimizer would
likely accelerate cases where the categorical variable deactivates
certain continuous components, as in type-I interaction and many
real-world problems. The lower-level optimizer also extends readily
to variants such as CMA-ES with margin, enabling simultaneous
handling of integer, continuous, and catego\-rical variables.
Finally, combining our method with lower-cost approaches such as
CatCMA or heuristics like early stopping or lower-level coevolution
may further alleviate the difficulty of bilevel optimization.

\begin{credits}
  \subsubsection{\ackname}
  This study was partly funded by JSPS KAKENHI 26K02993.
  \subsubsection{\discintname}
  The authors have no competing interests to declare that are relevant to the
  content of this article.

\end{credits}

\clearpage
\bibliographystyle{splncs04}
\bibliography{ppsn2026}

\clearpage
\appendix

\crefalias{section}{appendix}
\crefalias{subsection}{appendix}
\crefalias{subsubsection}{appendix}
\crefname{appendix}{Appendix}{Appendices}
\Crefname{appendix}{Appendix}{Appendices}

\section{Statistical Analysis of Algorithm Performance}\label{apdx:statistical}
\Cref{tab:fII,tab:fIIpIV_kappa1e2,tab:fIIpIV_kappa1e6,tab:fIII,tab:fIIIpIV_kappa1e2,tab:fIIIpIV_kappa1e6,tab:fII_10_10,tab:fIIpIV_kappa1e2_10_10,tab:fIIpIV_kappa1e6_10_10,tab:fIII_10_10,tab:fIIIpIV_kappa1e2_10_10,tab:fIIIpIV_kappa1e6_10_10} present the results of statistical hypothesis testing, comparing the proposed method against CatCMA and ICatCMA in terms of success rate, total number of function evaluations (FEs), and number of restarts, across all benchmark problems and settings reported in \Cref{ss:results}. Note that to accurately compare total computational cost, the statistics related to the number of total FEs and restarts reported in this section include all runs rather than only the successful ones, in contrast to \Cref{fig:5_5,fig:10_10} of the main paper. These data demonstrate the statistical significance of the following general trends discussed in \Cref{ss:results}:
\begin{enumerate}
    \item In both the $d_c=d_x=5$ and $d_c=d_x=10$ cases, for weak interaction regimes (small $a$), CatCMA outperforms IGBD.
    \item Otherwise, IGBD largely outperforms both CatCMA and ICatCMA.
    \item In the $d_c=d_x=5$ case, CatCMA is able to solve type-IV interaction using hundreds of restarts, suggesting a random search-like behavior.
\end{enumerate}
\begin{table}[h]
\centering
\caption{Comparison of the success rate, total number of function evaluations (FEs), and number of restarts of ICatCMA and CatCMA against the proposed method on the benchmark problem $\fII$ with $d_c = d_x = 5$ as a function of $a$. The main quantity and the parenthesized quantity represent the median and the IQR, respectively. The symbols $\uparrow$ and $\downarrow$ indicate that a quantity is statistically significantly higher or lower than the corresponding quantity for the proposed method under a Wilcoxon rank-sum test ($p<0.05$) with a Bonferroni correction accounting for all comparisons performed across \Cref{tab:fII,tab:fIIpIV_kappa1e2,tab:fIIpIV_kappa1e6,tab:fIII,tab:fIIIpIV_kappa1e2,tab:fIIIpIV_kappa1e6,tab:fII_10_10,tab:fIIpIV_kappa1e2_10_10,tab:fIIpIV_kappa1e6_10_10,tab:fIII_10_10,tab:fIIIpIV_kappa1e2_10_10,tab:fIIIpIV_kappa1e6_10_10}; the absence of a symbol indicates no statistically significant difference.}
\label{tab:fII}
\begin{tabular*}{\textwidth}{@{\extracolsep{\fill}}ccccc}
\hline
$a$ & & IGBD (ours) & CatCMA $\phantom{\downarrow}$ & ICatCMA $\phantom{\downarrow}$ \\ \hline
\multirow{3}{*}{0} & Success rate & \textcolor{blue}{1.00} & \textcolor{blue}{1.00 $\phantom{\downarrow}$} & \textcolor{blue}{1.00 $\phantom{\downarrow}$} \\
 & Total FEs & 5.26E+3 (1.14E+3) & \textcolor{blue}{6.95E+2 (7.75E+1) $\downarrow$} & 1.69E+4 (1.40E+3) $\uparrow$ \\
 & Restarts & 0 (0) & 0 (0) $\phantom{\downarrow}$ & 0 (0) $\phantom{\downarrow}$ \\ \hline
\multirow{3}{*}{1} & Success rate & \textcolor{blue}{1.00} & \textcolor{blue}{1.00 $\phantom{\downarrow}$} & \textcolor{blue}{1.00 $\phantom{\downarrow}$} \\
 & Total FEs & 9.51E+3 (5.28E+3) & \textcolor{blue}{7.20E+2 (1.10E+2) $\downarrow$} & 1.69E+4 (1.26E+3) $\uparrow$ \\
 & Restarts & 0 (0) & 0 (0) $\phantom{\downarrow}$ & 0 (0) $\phantom{\downarrow}$ \\ \hline
\multirow{3}{*}{2} & Success rate & \textcolor{blue}{1.00} & \textcolor{blue}{1.00 $\phantom{\downarrow}$} & \textcolor{blue}{1.00 $\phantom{\downarrow}$} \\
 & Total FEs & 9.66E+3 (4.08E+3) & 2.44E+4 (4.53E+4) $\phantom{\downarrow}$ & 1.74E+4 (1.39E+3) $\uparrow$ \\
 & Restarts & 0 (0) & 19 (36) $\uparrow$ & 0 (0) $\phantom{\downarrow}$ \\ \hline
\multirow{3}{*}{4} & Success rate & \textcolor{blue}{1.00} & 0.05 $\downarrow$ & \textcolor{blue}{1.00 $\phantom{\downarrow}$} \\
 & Total FEs & \textcolor{blue}{8.94E+3 (3.28E+3)} & 5.00E+5 (0.00E+0) $\uparrow$ & 1.86E+4 (8.54E+2) $\uparrow$ \\
 & Restarts & 0 (0) & 392 (2) $\uparrow$ & 0 (0) $\phantom{\downarrow}$ \\ \hline
\multirow{3}{*}{8} & Success rate & \textcolor{blue}{1.00} & 0.00 $\downarrow$ & \textcolor{blue}{1.00 $\phantom{\downarrow}$} \\
 & Total FEs & \textcolor{blue}{9.68E+3 (3.96E+3)} & 5.00E+5 (0.00E+0) $\uparrow$ & 1.97E+4 (1.60E+3) $\uparrow$ \\
 & Restarts & 0 (0) & 404 (2) $\uparrow$ & 0 (0) $\phantom{\downarrow}$ \\ \hline
\multirow{3}{*}{16} & Success rate & \textcolor{blue}{1.00} & 0.00 $\downarrow$ & \textcolor{blue}{1.00 $\phantom{\downarrow}$} \\
 & Total FEs & \textcolor{blue}{1.01E+4 (2.03E+3)} & 5.00E+5 (0.00E+0) $\uparrow$ & 2.09E+4 (2.00E+3) $\uparrow$ \\
 & Restarts & 0 (0) & 386 (3) $\uparrow$ & 0 (0) $\phantom{\downarrow}$ \\ \hline
\end{tabular*}
\end{table}

\begin{table}[t]
\centering
\caption{Comparison of the success rate, total number of function evaluations (FEs), and number of restarts of ICatCMA and CatCMA against the proposed method on the benchmark problem $\fIIpIV$ with $\kappa(\boldsymbol{A}) = 10^2$ and $d_c = d_x = 5$ as a function of $a$. The notation conventions for the median and IQR, as well as the statistical methodology and the meaning of the significance symbols $\uparrow$ and $\downarrow$, follow \Cref{tab:fII}.}
\label{tab:fIIpIV_kappa1e2}
\begin{tabular*}{\textwidth}{@{\extracolsep{\fill}}ccccc}
\hline
$a$ & & IGBD (ours) & CatCMA $\phantom{\downarrow}$ & ICatCMA $\phantom{\downarrow}$ \\ \hline
\multirow{3}{*}{0} & Success rate & \textcolor{blue}{1.00} & \textcolor{blue}{1.00 $\phantom{\downarrow}$} & \textcolor{blue}{1.00 $\phantom{\downarrow}$} \\
 & Total FEs & 6.54E+3 (7.52E+2) & \textcolor{blue}{1.18E+3 (2.90E+2) $\downarrow$} & 2.15E+4 (1.96E+3) $\uparrow$ \\
 & Restarts & 0 (0) & 0 (0) $\phantom{\downarrow}$ & 0 (0) $\phantom{\downarrow}$ \\ \hline
\multirow{3}{*}{1} & Success rate & \textcolor{blue}{1.00} & \textcolor{blue}{1.00 $\phantom{\downarrow}$} & \textcolor{blue}{1.00 $\phantom{\downarrow}$} \\
 & Total FEs & 1.03E+4 (3.84E+3) & \textcolor{blue}{2.79E+3 (2.87E+3) $\downarrow$} & 2.21E+4 (1.17E+3) $\uparrow$ \\
 & Restarts & 0 (0) & 1 (2) $\phantom{\downarrow}$ & 0 (0) $\phantom{\downarrow}$ \\ \hline
\multirow{3}{*}{2} & Success rate & \textcolor{blue}{1.00} & \textcolor{blue}{1.00 $\phantom{\downarrow}$} & \textcolor{blue}{1.00 $\phantom{\downarrow}$} \\
 & Total FEs & 1.04E+4 (3.43E+3) & 1.75E+4 (3.64E+4) $\phantom{\downarrow}$ & 2.45E+4 (2.30E+4) $\uparrow$ \\
 & Restarts & 0 (0) & 9 (22) $\uparrow$ & 0 (1) $\phantom{\downarrow}$ \\ \hline
\multirow{3}{*}{4} & Success rate & \textcolor{blue}{1.00} & \textcolor{blue}{1.00 $\phantom{\downarrow}$} & \textcolor{blue}{1.00 $\phantom{\downarrow}$} \\
 & Total FEs & \textcolor{blue}{9.75E+3 (5.74E+3)} & 3.77E+4 (7.91E+4) $\uparrow$ & 6.19E+4 (8.62E+4) $\uparrow$ \\
 & Restarts & 0 (0) & 23 (50) $\uparrow$ & 2 (3) $\phantom{\downarrow}$ \\ \hline
\multirow{3}{*}{8} & Success rate & \textcolor{blue}{1.00} & \textcolor{blue}{1.00 $\phantom{\downarrow}$} & 0.85 $\phantom{\downarrow}$ \\
 & Total FEs & \textcolor{blue}{1.05E+4 (3.54E+3)} & 5.00E+4 (5.07E+4) $\uparrow$ & 2.66E+5 (3.01E+5) $\uparrow$ \\
 & Restarts & 0 (0) & 30 (31) $\uparrow$ & 10 (12) $\uparrow$ \\ \hline
\multirow{3}{*}{16} & Success rate & \textcolor{blue}{1.00} & \textcolor{blue}{1.00 $\phantom{\downarrow}$} & 0.25 $\downarrow$ \\
 & Total FEs & \textcolor{blue}{1.22E+4 (5.33E+3)} & 3.89E+4 (3.61E+4) $\uparrow$ & 5.00E+5 (6.71E+3) $\uparrow$ \\
 & Restarts & 0 (0) & 22 (22) $\uparrow$ & 18 (0) $\uparrow$ \\ \hline
\end{tabular*}
\end{table}

\begin{table}[t]
\centering
\caption{Comparison of the success rate, total number of function evaluations (FEs), and number of restarts of ICatCMA and CatCMA against the proposed method on the benchmark problem $\fIIpIV$ with $\kappa(\boldsymbol{A}) = 10^6$ and $d_c = d_x = 5$ as a function of $a$. The notation conventions for the median and IQR, as well as the statistical methodology and the meaning of the significance symbols $\uparrow$ and $\downarrow$, follow \Cref{tab:fII}.}
\label{tab:fIIpIV_kappa1e6}
\begin{tabular*}{\textwidth}{@{\extracolsep{\fill}}ccccc}
\hline
$a$ & & IGBD (ours) & CatCMA $\phantom{\downarrow}$ & ICatCMA $\phantom{\downarrow}$ \\ \hline
\multirow{3}{*}{0} & Success rate & \textcolor{blue}{1.00} & \textcolor{blue}{1.00 $\phantom{\downarrow}$} & 0.05 $\downarrow$ \\
 & Total FEs & 2.66E+4 (4.61E+4) & \textcolor{blue}{3.59E+3 (3.70E+2) $\downarrow$} & 5.00E+5 (0.00E+0) $\uparrow$ \\
 & Restarts & 0 (0) & 0 (0) $\phantom{\downarrow}$ & 12 (0) $\uparrow$ \\ \hline
\multirow{3}{*}{1} & Success rate & \textcolor{blue}{1.00} & \textcolor{blue}{1.00 $\phantom{\downarrow}$} & 0.15 $\downarrow$ \\
 & Total FEs & 2.63E+4 (3.07E+4) & 3.23E+4 (3.46E+4) $\phantom{\downarrow}$ & 5.00E+5 (0.00E+0) $\uparrow$ \\
 & Restarts & 0 (0) & 10 (11) $\uparrow$ & 12 (0) $\uparrow$ \\ \hline
\multirow{3}{*}{2} & Success rate & \textcolor{blue}{1.00} & \textcolor{blue}{1.00 $\phantom{\downarrow}$} & 0.10 $\downarrow$ \\
 & Total FEs & 4.45E+4 (2.39E+4) & 1.82E+4 (1.84E+4) $\phantom{\downarrow}$ & 5.00E+5 (0.00E+0) $\uparrow$ \\
 & Restarts & 0 (0) & 5 (6) $\uparrow$ & 12 (1) $\uparrow$ \\ \hline
\multirow{3}{*}{4} & Success rate & \textcolor{blue}{1.00} & \textcolor{blue}{1.00 $\phantom{\downarrow}$} & 0.05 $\downarrow$ \\
 & Total FEs & 4.13E+4 (2.01E+4) & 7.03E+4 (5.74E+4) $\phantom{\downarrow}$ & 5.00E+5 (0.00E+0) $\uparrow$ \\
 & Restarts & 0 (0) & 22 (19) $\uparrow$ & 11 (1) $\uparrow$ \\ \hline
\multirow{3}{*}{8} & Success rate & \textcolor{blue}{1.00} & \textcolor{blue}{1.00 $\phantom{\downarrow}$} & 0.10 $\downarrow$ \\
 & Total FEs & 3.64E+4 (2.00E+4) & 3.48E+4 (4.54E+4) $\phantom{\downarrow}$ & 5.00E+5 (0.00E+0) $\uparrow$ \\
 & Restarts & 0 (0) & 10 (14) $\uparrow$ & 11 (0) $\uparrow$ \\ \hline
\multirow{3}{*}{16} & Success rate & \textcolor{blue}{1.00} & \textcolor{blue}{1.00 $\phantom{\downarrow}$} & 0.05 $\downarrow$ \\
 & Total FEs & 4.92E+4 (3.57E+4) & 7.73E+4 (1.32E+5) $\phantom{\downarrow}$ & 5.00E+5 (0.00E+0) $\uparrow$ \\
 & Restarts & 0 (0) & 24 (41) $\uparrow$ & 11 (0) $\uparrow$ \\ \hline
\end{tabular*}
\end{table}

\begin{table}[t]
\centering
\caption{Comparison of the success rate, total number of function evaluations (FEs), and number of restarts of ICatCMA and CatCMA against the proposed method on the benchmark problem $\fIII$ with $d_c = d_x = 5$ as a function of $a$. The notation conventions for the median and IQR, as well as the statistical methodology and the meaning of the significance symbols $\uparrow$ and $\downarrow$, follow \Cref{tab:fII}.}
\label{tab:fIII}
\begin{tabular*}{\textwidth}{@{\extracolsep{\fill}}ccccc}
\hline
$a$ & & IGBD (ours) & CatCMA $\phantom{\downarrow}$ & ICatCMA $\phantom{\downarrow}$ \\ \hline
\multirow{3}{*}{0} & Success rate & \textcolor{blue}{1.00} & \textcolor{blue}{1.00 $\phantom{\downarrow}$} & \textcolor{blue}{1.00 $\phantom{\downarrow}$} \\
 & Total FEs & 1.01E+4 (4.99E+3) & \textcolor{blue}{2.02E+3 (3.19E+3) $\downarrow$} & 1.99E+4 (2.11E+4) $\uparrow$ \\
 & Restarts & 0 (0) & 1 (3) $\phantom{\downarrow}$ & 0 (1) $\phantom{\downarrow}$ \\ \hline
\multirow{3}{*}{1} & Success rate & \textcolor{blue}{1.00} & 0.20 $\downarrow$ & \textcolor{blue}{1.00 $\phantom{\downarrow}$} \\
 & Total FEs & 1.36E+4 (3.28E+4) & 5.00E+5 (0.00E+0) $\uparrow$ & 2.01E+4 (2.46E+4) $\phantom{\downarrow}$ \\
 & Restarts & 0 (0) & 404 (6) $\uparrow$ & 0 (1) $\phantom{\downarrow}$ \\ \hline
\multirow{3}{*}{2} & Success rate & \textcolor{blue}{1.00} & 0.00 $\downarrow$ & 0.95 $\phantom{\downarrow}$ \\
 & Total FEs & 3.51E+4 (5.11E+4) & 5.00E+5 (0.00E+0) $\uparrow$ & 1.72E+5 (2.32E+5) $\phantom{\downarrow}$ \\
 & Restarts & 0 (0) & 394 (6) $\uparrow$ & 6 (10) $\uparrow$ \\ \hline
\multirow{3}{*}{4} & Success rate & \textcolor{blue}{1.00} & 0.00 $\downarrow$ & 0.95 $\phantom{\downarrow}$ \\
 & Total FEs & \textcolor{blue}{1.93E+4 (1.62E+4)} & 5.00E+5 (0.00E+0) $\uparrow$ & 1.02E+5 (8.93E+4) $\uparrow$ \\
 & Restarts & 0 (0) & 412 (6) $\uparrow$ & 3 (3) $\uparrow$ \\ \hline
\multirow{3}{*}{8} & Success rate & \textcolor{blue}{1.00} & 0.00 $\downarrow$ & 0.75 $\phantom{\downarrow}$ \\
 & Total FEs & \textcolor{blue}{1.79E+4 (1.58E+4)} & 5.00E+5 (0.00E+0) $\uparrow$ & 2.86E+5 (3.16E+5) $\uparrow$ \\
 & Restarts & 0 (0) & 538 (15) $\uparrow$ & 13 (14) $\uparrow$ \\ \hline
\multirow{3}{*}{16} & Success rate & 0.95 & 0.00 $\downarrow$ & 0.75 $\phantom{\downarrow}$ \\
 & Total FEs & \textcolor{blue}{2.02E+4 (2.90E+4)} & 5.00E+5 (0.00E+0) $\uparrow$ & 2.99E+5 (2.82E+5) $\uparrow$ \\
 & Restarts & 0 (0) & 662 (21) $\uparrow$ & 14 (14) $\uparrow$ \\ \hline
\end{tabular*}
\end{table}

\begin{table}[t]
\centering
\caption{Comparison of the success rate, total number of function evaluations (FEs), and number of restarts of ICatCMA and CatCMA against the proposed method on the benchmark problem $\fIIIpIV$ with $\kappa(\boldsymbol{A}) = 10^2$ and $d_c = d_x = 5$ as a function of $a$. The notation conventions for the median and IQR, as well as the statistical methodology and the meaning of the significance symbols $\uparrow$ and $\downarrow$, follow \Cref{tab:fII}.}
\label{tab:fIIIpIV_kappa1e2}
\begin{tabular*}{\textwidth}{@{\extracolsep{\fill}}ccccc}
\hline
$a$ & & IGBD (ours) & CatCMA $\phantom{\downarrow}$ & ICatCMA $\phantom{\downarrow}$ \\ \hline
\multirow{3}{*}{0} & Success rate & \textcolor{blue}{1.00} & \textcolor{blue}{1.00 $\phantom{\downarrow}$} & 0.05 $\downarrow$ \\
 & Total FEs & 1.26E+4 (9.44E+3) & 2.33E+4 (2.32E+4) $\phantom{\downarrow}$ & 5.00E+5 (0.00E+0) $\uparrow$ \\
 & Restarts & 0 (0) & 15 (15) $\uparrow$ & 15 (0) $\uparrow$ \\ \hline
\multirow{3}{*}{1} & Success rate & 0.95 & \textcolor{blue}{1.00 $\phantom{\downarrow}$} & 0.00 $\downarrow$ \\
 & Total FEs & 1.19E+4 (1.15E+4) & 1.94E+4 (2.15E+4) $\phantom{\downarrow}$ & 5.00E+5 (0.00E+0) $\uparrow$ \\
 & Restarts & 0 (0) & 13 (14) $\uparrow$ & 15 (0) $\uparrow$ \\ \hline
\multirow{3}{*}{2} & Success rate & \textcolor{blue}{1.00} & \textcolor{blue}{1.00 $\phantom{\downarrow}$} & 0.00 $\downarrow$ \\
 & Total FEs & 1.83E+4 (3.30E+4) & 4.19E+4 (1.08E+5) $\phantom{\downarrow}$ & 5.00E+5 (0.00E+0) $\uparrow$ \\
 & Restarts & 0 (0) & 28 (70) $\uparrow$ & 15 (0) $\uparrow$ \\ \hline
\multirow{3}{*}{4} & Success rate & \textcolor{blue}{1.00} & 0.90 $\phantom{\downarrow}$ & 0.00 $\downarrow$ \\
 & Total FEs & \textcolor{blue}{1.11E+4 (4.54E+3)} & 8.83E+4 (1.66E+5) $\uparrow$ & 5.00E+5 (0.00E+0) $\uparrow$ \\
 & Restarts & 0 (0) & 64 (119) $\uparrow$ & 15 (1) $\uparrow$ \\ \hline
\multirow{3}{*}{8} & Success rate & \textcolor{blue}{1.00} & \textcolor{blue}{1.00 $\phantom{\downarrow}$} & 0.00 $\downarrow$ \\
 & Total FEs & 1.32E+4 (8.24E+3) & 4.51E+4 (9.72E+4) $\phantom{\downarrow}$ & 5.00E+5 (0.00E+0) $\uparrow$ \\
 & Restarts & 0 (0) & 31 (68) $\uparrow$ & 15 (1) $\uparrow$ \\ \hline
\multirow{3}{*}{16} & Success rate & \textcolor{blue}{1.00} & \textcolor{blue}{1.00 $\phantom{\downarrow}$} & 0.00 $\downarrow$ \\
 & Total FEs & 1.53E+4 (1.32E+4) & 1.81E+4 (2.72E+4) $\phantom{\downarrow}$ & 5.00E+5 (0.00E+0) $\uparrow$ \\
 & Restarts & 0 (0) & 15 (24) $\uparrow$ & 31 (1) $\uparrow$ \\ \hline
\end{tabular*}
\end{table}

\begin{table}[t]
\centering
\caption{Comparison of the success rate, total number of function evaluations (FEs), and number of restarts of ICatCMA and CatCMA against the proposed method on the benchmark problem $\fIIIpIV$ with $\kappa(\boldsymbol{A}) = 10^6$ and $d_c = d_x = 5$ as a function of $a$. The notation conventions for the median and IQR, as well as the statistical methodology and the meaning of the significance symbols $\uparrow$ and $\downarrow$, follow \Cref{tab:fII}.}
\label{tab:fIIIpIV_kappa1e6}
\begin{tabular*}{\textwidth}{@{\extracolsep{\fill}}ccccc}
\hline
$a$ & & IGBD (ours) & CatCMA $\phantom{\downarrow}$ & ICatCMA $\phantom{\downarrow}$ \\ \hline
\multirow{3}{*}{0} & Success rate & \textcolor{blue}{1.00} & \textcolor{blue}{1.00 $\phantom{\downarrow}$} & 0.00 $\downarrow$ \\
 & Total FEs & 1.71E+4 (7.50E+3) & 6.78E+4 (7.45E+4) $\phantom{\downarrow}$ & 5.00E+5 (0.00E+0) $\uparrow$ \\
 & Restarts & 0 (0) & 30 (33) $\uparrow$ & 11 (1) $\uparrow$ \\ \hline
\multirow{3}{*}{1} & Success rate & \textcolor{blue}{1.00} & \textcolor{blue}{1.00 $\phantom{\downarrow}$} & 0.00 $\downarrow$ \\
 & Total FEs & 1.96E+4 (1.52E+4) & 3.26E+4 (5.11E+4) $\phantom{\downarrow}$ & 5.00E+5 (0.00E+0) $\uparrow$ \\
 & Restarts & 0 (0) & 13 (20) $\uparrow$ & 11 (0) $\uparrow$ \\ \hline
\multirow{3}{*}{2} & Success rate & \textcolor{blue}{1.00} & \textcolor{blue}{1.00 $\phantom{\downarrow}$} & 0.00 $\downarrow$ \\
 & Total FEs & 2.54E+4 (2.94E+4) & 9.36E+3 (2.07E+4) $\phantom{\downarrow}$ & 5.00E+5 (0.00E+0) $\uparrow$ \\
 & Restarts & 0 (0) & 3 (9) $\uparrow$ & 11 (0) $\uparrow$ \\ \hline
\multirow{3}{*}{4} & Success rate & \textcolor{blue}{1.00} & \textcolor{blue}{1.00 $\phantom{\downarrow}$} & 0.00 $\downarrow$ \\
 & Total FEs & 1.86E+4 (8.11E+3) & 5.88E+4 (6.29E+4) $\phantom{\downarrow}$ & 5.00E+5 (0.00E+0) $\uparrow$ \\
 & Restarts & 0 (0) & 28 (31) $\uparrow$ & 10 (1) $\uparrow$ \\ \hline
\multirow{3}{*}{8} & Success rate & \textcolor{blue}{1.00} & \textcolor{blue}{1.00 $\phantom{\downarrow}$} & 0.00 $\downarrow$ \\
 & Total FEs & 1.84E+4 (5.05E+3) & 2.68E+4 (3.86E+4) $\phantom{\downarrow}$ & 5.00E+5 (0.00E+0) $\uparrow$ \\
 & Restarts & 0 (0) & 10 (18) $\uparrow$ & 10 (1) $\uparrow$ \\ \hline
\multirow{3}{*}{16} & Success rate & \textcolor{blue}{1.00} & \textcolor{blue}{1.00 $\phantom{\downarrow}$} & 0.00 $\downarrow$ \\
 & Total FEs & 1.97E+4 (6.37E+3) & 5.98E+4 (4.73E+4) $\phantom{\downarrow}$ & 5.00E+5 (0.00E+0) $\uparrow$ \\
 & Restarts & 0 (0) & 27 (22) $\uparrow$ & 10 (2) $\uparrow$ \\ \hline
\end{tabular*}
\end{table}

\begin{table}[t]
\centering
\caption{Comparison of the success rate, total number of function evaluations (FEs), and number of restarts of ICatCMA and CatCMA against the proposed method on the benchmark problem $\fII$ with $d_c = d_x = 10$ as a function of $a$. The notation conventions for the median and IQR, as well as the statistical methodology and the meaning of the significance symbols $\uparrow$ and $\downarrow$, follow \Cref{tab:fII}.}
\label{tab:fII_10_10}
\begin{tabular*}{\textwidth}{@{\extracolsep{\fill}}ccccc}
\hline
$a$ & & IGBD (ours) & CatCMA $\phantom{\downarrow}$ & ICatCMA $\phantom{\downarrow}$ \\ \hline
\multirow{3}{*}{0} & Success rate & \textcolor{blue}{1.00} & \textcolor{blue}{1.00 $\phantom{\downarrow}$} & \textcolor{blue}{1.00 $\phantom{\downarrow}$} \\
 & Total FEs & 1.20E+4 (2.12E+3) & \textcolor{blue}{1.34E+3 (1.68E+2) $\downarrow$} & 7.41E+4 (8.80E+3) $\uparrow$ \\
 & Restarts & 0 (0) & 0 (0) $\phantom{\downarrow}$ & 0 (0) $\phantom{\downarrow}$ \\ \hline
\multirow{3}{*}{1} & Success rate & \textcolor{blue}{1.00} & \textcolor{blue}{1.00 $\phantom{\downarrow}$} & \textcolor{blue}{1.00 $\phantom{\downarrow}$} \\
 & Total FEs & 2.53E+4 (1.13E+4) & \textcolor{blue}{1.44E+3 (1.47E+2) $\downarrow$} & 7.05E+4 (4.95E+3) $\uparrow$ \\
 & Restarts & 0 (0) & 0 (0) $\phantom{\downarrow}$ & 0 (0) $\phantom{\downarrow}$ \\ \hline
\multirow{3}{*}{2} & Success rate & \textcolor{blue}{1.00} & \textcolor{blue}{1.00 $\phantom{\downarrow}$} & \textcolor{blue}{1.00 $\phantom{\downarrow}$} \\
 & Total FEs & 2.67E+4 (6.01E+3) & \textcolor{blue}{1.59E+3 (2.25E+2) $\downarrow$} & 7.20E+4 (5.71E+3) $\uparrow$ \\
 & Restarts & 0 (0) & 0 (0) $\phantom{\downarrow}$ & 0 (0) $\phantom{\downarrow}$ \\ \hline
\multirow{3}{*}{4} & Success rate & \textcolor{blue}{1.00} & \textcolor{blue}{1.00 $\phantom{\downarrow}$} & \textcolor{blue}{1.00 $\phantom{\downarrow}$} \\
 & Total FEs & 2.80E+4 (9.39E+3) & 7.79E+4 (2.06E+5) $\phantom{\downarrow}$ & 8.12E+4 (5.72E+3) $\uparrow$ \\
 & Restarts & 0 (0) & 32 (88) $\uparrow$ & 0 (0) $\phantom{\downarrow}$ \\ \hline
\multirow{3}{*}{8} & Success rate & \textcolor{blue}{1.00} & 0.00 $\downarrow$ & \textcolor{blue}{1.00 $\phantom{\downarrow}$} \\
 & Total FEs & \textcolor{blue}{2.78E+4 (1.31E+4)} & 2.00E+6 (0.00E+0) $\uparrow$ & 8.45E+4 (5.53E+3) $\uparrow$ \\
 & Restarts & 0 (0) & 927 (4) $\uparrow$ & 0 (0) $\phantom{\downarrow}$ \\ \hline
\multirow{3}{*}{16} & Success rate & \textcolor{blue}{1.00} & 0.00 $\downarrow$ & \textcolor{blue}{1.00 $\phantom{\downarrow}$} \\
 & Total FEs & \textcolor{blue}{2.97E+4 (8.73E+3)} & 2.00E+6 (0.00E+0) $\uparrow$ & 8.88E+4 (9.94E+3) $\uparrow$ \\
 & Restarts & 0 (0) & 926 (5) $\uparrow$ & 0 (0) $\phantom{\downarrow}$ \\ \hline
\end{tabular*}
\end{table}

\begin{table}[t]
\centering
\caption{Comparison of the success rate, total number of function evaluations (FEs), and number of restarts of ICatCMA and CatCMA against the proposed method on the benchmark problem $\fIIpIV$ with $\kappa(\boldsymbol{A}) = 10^2$ and $d_c = d_x = 10$ as a function of $a$. The notation conventions for the median and IQR, as well as the statistical methodology and the meaning of the significance symbols $\uparrow$ and $\downarrow$, follow \Cref{tab:fII}.}
\label{tab:fIIpIV_kappa1e2_10_10}
\begin{tabular*}{\textwidth}{@{\extracolsep{\fill}}ccccc}
\hline
$a$ & & IGBD (ours) & CatCMA $\phantom{\downarrow}$ & ICatCMA $\phantom{\downarrow}$ \\ \hline
\multirow{3}{*}{0} & Success rate & \textcolor{blue}{1.00} & \textcolor{blue}{1.00 $\phantom{\downarrow}$} & \textcolor{blue}{1.00 $\phantom{\downarrow}$} \\
 & Total FEs & 1.72E+4 (3.17E+3) & \textcolor{blue}{3.02E+3 (3.21E+2) $\downarrow$} & 9.16E+4 (9.93E+3) $\uparrow$ \\
 & Restarts & 0 (0) & 0 (0) $\phantom{\downarrow}$ & 0 (0) $\phantom{\downarrow}$ \\ \hline
\multirow{3}{*}{1} & Success rate & \textcolor{blue}{1.00} & \textcolor{blue}{1.00 $\phantom{\downarrow}$} & \textcolor{blue}{1.00 $\phantom{\downarrow}$} \\
 & Total FEs & 3.54E+4 (9.76E+3) & 5.03E+4 (1.03E+5) $\phantom{\downarrow}$ & 9.36E+4 (5.17E+3) $\uparrow$ \\
 & Restarts & 0 (0) & 11 (26) $\uparrow$ & 0 (0) $\phantom{\downarrow}$ \\ \hline
\multirow{3}{*}{2} & Success rate & \textcolor{blue}{1.00} & 0.00 $\downarrow$ & \textcolor{blue}{1.00 $\phantom{\downarrow}$} \\
 & Total FEs & \textcolor{blue}{3.38E+4 (8.54E+3)} & 2.00E+6 (0.00E+0) $\uparrow$ & 1.04E+5 (4.86E+3) $\uparrow$ \\
 & Restarts & 0 (0) & 606 (6) $\uparrow$ & 0 (0) $\phantom{\downarrow}$ \\ \hline
\multirow{3}{*}{4} & Success rate & \textcolor{blue}{1.00} & 0.00 $\downarrow$ & \textcolor{blue}{1.00 $\phantom{\downarrow}$} \\
 & Total FEs & \textcolor{blue}{3.98E+4 (1.37E+4)} & 2.00E+6 (0.00E+0) $\uparrow$ & 2.70E+5 (7.03E+5) $\uparrow$ \\
 & Restarts & 0 (0) & 641 (2) $\uparrow$ & 2 (6) $\phantom{\downarrow}$ \\ \hline
\multirow{3}{*}{8} & Success rate & \textcolor{blue}{1.00} & 0.00 $\downarrow$ & 0.20 $\phantom{\downarrow}$ \\
 & Total FEs & \textcolor{blue}{6.98E+4 (7.67E+4)} & 2.00E+6 (0.00E+0) $\uparrow$ & 2.00E+6 (0.00E+0) $\uparrow$ \\
 & Restarts & 0 (0) & 618 (1) $\uparrow$ & 17 (0) $\uparrow$ \\ \hline
\multirow{3}{*}{16} & Success rate & \textcolor{blue}{1.00} & 0.00 $\downarrow$ & 0.00 $\downarrow$ \\
 & Total FEs & \textcolor{blue}{3.15E+5 (3.97E+5)} & 2.00E+6 (0.00E+0) $\uparrow$ & 2.00E+6 (0.00E+0) $\uparrow$ \\
 & Restarts & 0 (0) & 606 (3) $\uparrow$ & 16 (0) $\uparrow$ \\ \hline
\end{tabular*}
\end{table}

\begin{table}[t]
\centering
\caption{Comparison of the success rate, total number of function evaluations (FEs), and number of restarts of ICatCMA and CatCMA against the proposed method on the benchmark problem $\fIIpIV$ with $\kappa(\boldsymbol{A}) = 10^6$ and $d_c = d_x = 10$ as a function of $a$. The notation conventions for the median and IQR, as well as the statistical methodology and the meaning of the significance symbols $\uparrow$ and $\downarrow$, follow \Cref{tab:fII}.}
\label{tab:fIIpIV_kappa1e6_10_10}
\begin{tabular*}{\textwidth}{@{\extracolsep{\fill}}ccccc}
\hline
$a$ & & IGBD (ours) & CatCMA $\phantom{\downarrow}$ & ICatCMA $\phantom{\downarrow}$ \\ \hline
\multirow{3}{*}{0} & Success rate & 0.15 & \textcolor{blue}{1.00 $\uparrow$} & 0.00 $\phantom{\downarrow}$ \\
 & Total FEs & 2.00E+6 (3.47E+3) & \textcolor{blue}{2.37E+4 (2.25E+4) $\downarrow$} & 2.00E+6 (0.00E+0) $\phantom{\downarrow}$ \\
 & Restarts & 2 (1) & 1 (2) $\phantom{\downarrow}$ & 8 (1) $\uparrow$ \\ \hline
\multirow{3}{*}{1} & Success rate & 0.65 & 0.00 $\phantom{\downarrow}$ & 0.00 $\phantom{\downarrow}$ \\
 & Total FEs & 1.77E+6 (1.23E+6) & 2.00E+6 (0.00E+0) $\phantom{\downarrow}$ & 2.00E+6 (0.00E+0) $\phantom{\downarrow}$ \\
 & Restarts & 0 (2) & 251 (1) $\uparrow$ & 7 (1) $\uparrow$ \\ \hline
\multirow{3}{*}{2} & Success rate & 0.80 & 0.00 $\downarrow$ & 0.00 $\phantom{\downarrow}$ \\
 & Total FEs & 1.13E+6 (9.73E+5) & 2.00E+6 (0.00E+0) $\phantom{\downarrow}$ & 2.00E+6 (0.00E+0) $\phantom{\downarrow}$ \\
 & Restarts & 0 (1) & 249 (2) $\uparrow$ & 7 (0) $\uparrow$ \\ \hline
\multirow{3}{*}{4} & Success rate & 0.85 & 0.00 $\downarrow$ & 0.00 $\phantom{\downarrow}$ \\
 & Total FEs & 9.06E+5 (7.81E+5) & 2.00E+6 (0.00E+0) $\phantom{\downarrow}$ & 2.00E+6 (0.00E+0) $\phantom{\downarrow}$ \\
 & Restarts & 0 (0) & 245 (2) $\uparrow$ & 7 (0) $\uparrow$ \\ \hline
\multirow{3}{*}{8} & Success rate & 0.80 & 0.00 $\downarrow$ & 0.00 $\phantom{\downarrow}$ \\
 & Total FEs & 8.74E+5 (9.77E+5) & 2.00E+6 (0.00E+0) $\phantom{\downarrow}$ & 2.00E+6 (0.00E+0) $\phantom{\downarrow}$ \\
 & Restarts & 0 (0) & 240 (1) $\uparrow$ & 7 (0) $\uparrow$ \\ \hline
\multirow{3}{*}{16} & Success rate & 0.60 & 0.00 $\phantom{\downarrow}$ & 0.00 $\phantom{\downarrow}$ \\
 & Total FEs & 1.30E+6 (1.02E+6) & 2.00E+6 (0.00E+0) $\phantom{\downarrow}$ & 2.00E+6 (0.00E+0) $\phantom{\downarrow}$ \\
 & Restarts & 0 (0) & 235 (1) $\uparrow$ & 7 (0) $\uparrow$ \\ \hline
\end{tabular*}
\end{table}

\begin{table}[t]
\centering
\caption{Comparison of the success rate, total number of function evaluations (FEs), and number of restarts of ICatCMA and CatCMA against the proposed method on the benchmark problem $\fIII$ with $d_c = d_x = 10$ as a function of $a$. The notation conventions for the median and IQR, as well as the statistical methodology and the meaning of the significance symbols $\uparrow$ and $\downarrow$, follow \Cref{tab:fII}.}
\label{tab:fIII_10_10}
\begin{tabular*}{\textwidth}{@{\extracolsep{\fill}}ccccc}
\hline
$a$ & & IGBD (ours) & CatCMA $\phantom{\downarrow}$ & ICatCMA $\phantom{\downarrow}$ \\ \hline
\multirow{3}{*}{0} & Success rate & \textcolor{blue}{1.00} & \textcolor{blue}{1.00 $\phantom{\downarrow}$} & \textcolor{blue}{1.00 $\phantom{\downarrow}$} \\
 & Total FEs & 3.42E+4 (2.00E+4) & 1.89E+4 (4.55E+4) $\phantom{\downarrow}$ & 3.61E+5 (4.08E+5) $\uparrow$ \\
 & Restarts & 0 (0) & 8 (22) $\uparrow$ & 4 (5) $\phantom{\downarrow}$ \\ \hline
\multirow{3}{*}{1} & Success rate & \textcolor{blue}{1.00} & 0.00 $\downarrow$ & \textcolor{blue}{1.00 $\phantom{\downarrow}$} \\
 & Total FEs & 7.22E+4 (1.11E+5) & 2.00E+6 (0.00E+0) $\uparrow$ & 2.48E+5 (2.37E+5) $\phantom{\downarrow}$ \\
 & Restarts & 0 (1) & 906 (8) $\uparrow$ & 2 (2) $\phantom{\downarrow}$ \\ \hline
\multirow{3}{*}{2} & Success rate & \textcolor{blue}{1.00} & 0.00 $\downarrow$ & \textcolor{blue}{1.00 $\phantom{\downarrow}$} \\
 & Total FEs & 7.87E+4 (9.68E+4) & 2.00E+6 (0.00E+0) $\uparrow$ & 3.48E+5 (4.76E+5) $\phantom{\downarrow}$ \\
 & Restarts & 0 (0) & 939 (10) $\uparrow$ & 3 (6) $\phantom{\downarrow}$ \\ \hline
\multirow{3}{*}{4} & Success rate & \textcolor{blue}{1.00} & 0.00 $\downarrow$ & \textcolor{blue}{1.00 $\phantom{\downarrow}$} \\
 & Total FEs & 1.19E+5 (2.16E+5) & 2.00E+6 (0.00E+0) $\uparrow$ & 2.75E+5 (3.85E+5) $\phantom{\downarrow}$ \\
 & Restarts & 0 (0) & 1915 (56) $\uparrow$ & 2 (4) $\phantom{\downarrow}$ \\ \hline
\multirow{3}{*}{8} & Success rate & 0.85 & 0.00 $\downarrow$ & 0.70 $\phantom{\downarrow}$ \\
 & Total FEs & 2.70E+5 (8.85E+5) & 2.00E+6 (0.00E+0) $\phantom{\downarrow}$ & 1.13E+6 (1.36E+6) $\phantom{\downarrow}$ \\
 & Restarts & 0 (1) & 2006 (70) $\uparrow$ & 10 (14) $\uparrow$ \\ \hline
\multirow{3}{*}{16} & Success rate & 0.95 & 0.00 $\downarrow$ & 0.80 $\phantom{\downarrow}$ \\
 & Total FEs & 2.59E+5 (5.52E+5) & 2.00E+6 (0.00E+0) $\uparrow$ & 1.03E+6 (1.12E+6) $\phantom{\downarrow}$ \\
 & Restarts & 0 (1) & 2338 (37) $\uparrow$ & 8 (10) $\phantom{\downarrow}$ \\ \hline
\end{tabular*}
\end{table}

\begin{table}[t]
\centering
\caption{Comparison of the success rate, total number of function evaluations (FEs), and number of restarts of ICatCMA and CatCMA against the proposed method on the benchmark problem $\fIIIpIV$ with $\kappa(\boldsymbol{A}) = 10^2$ and $d_c = d_x = 10$ as a function of $a$. The notation conventions for the median and IQR, as well as the statistical methodology and the meaning of the significance symbols $\uparrow$ and $\downarrow$, follow \Cref{tab:fII}.}
\label{tab:fIIIpIV_kappa1e2_10_10}
\begin{tabular*}{\textwidth}{@{\extracolsep{\fill}}ccccc}
\hline
$a$ & & IGBD (ours) & CatCMA $\phantom{\downarrow}$ & ICatCMA $\phantom{\downarrow}$ \\ \hline
\multirow{3}{*}{0} & Success rate & \textcolor{blue}{1.00} & 0.05 $\downarrow$ & 0.00 $\downarrow$ \\
 & Total FEs & \textcolor{blue}{7.04E+4 (7.02E+4)} & 2.00E+6 (0.00E+0) $\uparrow$ & 2.00E+6 (0.00E+0) $\uparrow$ \\
 & Restarts & 0 (0) & 778 (6) $\uparrow$ & 14 (1) $\uparrow$ \\ \hline
\multirow{3}{*}{1} & Success rate & \textcolor{blue}{1.00} & 0.00 $\downarrow$ & 0.00 $\downarrow$ \\
 & Total FEs & \textcolor{blue}{1.42E+5 (1.65E+5)} & 2.00E+6 (0.00E+0) $\uparrow$ & 2.00E+6 (0.00E+0) $\uparrow$ \\
 & Restarts & 0 (0) & 792 (8) $\uparrow$ & 14 (1) $\uparrow$ \\ \hline
\multirow{3}{*}{2} & Success rate & \textcolor{blue}{1.00} & 0.00 $\downarrow$ & 0.00 $\downarrow$ \\
 & Total FEs & \textcolor{blue}{5.85E+4 (1.12E+5)} & 2.00E+6 (0.00E+0) $\uparrow$ & 2.00E+6 (0.00E+0) $\uparrow$ \\
 & Restarts & 0 (0) & 772 (4) $\uparrow$ & 13 (1) $\uparrow$ \\ \hline
\multirow{3}{*}{4} & Success rate & \textcolor{blue}{1.00} & 0.00 $\downarrow$ & 0.00 $\downarrow$ \\
 & Total FEs & \textcolor{blue}{9.35E+4 (6.93E+4)} & 2.00E+6 (0.00E+0) $\uparrow$ & 2.00E+6 (0.00E+0) $\uparrow$ \\
 & Restarts & 0 (0) & 738 (6) $\uparrow$ & 11 (0) $\uparrow$ \\ \hline
\multirow{3}{*}{8} & Success rate & \textcolor{blue}{1.00} & 0.00 $\downarrow$ & 0.00 $\downarrow$ \\
 & Total FEs & \textcolor{blue}{2.02E+5 (2.13E+5)} & 2.00E+6 (0.00E+0) $\uparrow$ & 2.00E+6 (0.00E+0) $\uparrow$ \\
 & Restarts & 0 (0) & 674 (9) $\uparrow$ & 12 (2) $\uparrow$ \\ \hline
\multirow{3}{*}{16} & Success rate & \textcolor{blue}{1.00} & 0.00 $\downarrow$ & 0.10 $\downarrow$ \\
 & Total FEs & \textcolor{blue}{3.77E+5 (7.73E+5)} & 2.00E+6 (0.00E+0) $\uparrow$ & 2.00E+6 (0.00E+0) $\uparrow$ \\
 & Restarts & 0 (1) & 1284 (28) $\uparrow$ & 16 (2) $\uparrow$ \\ \hline
\end{tabular*}
\end{table}

\begin{table}[t]
\centering
\caption{Comparison of the success rate, total number of function evaluations (FEs), and number of restarts of ICatCMA and CatCMA against the proposed method on the benchmark problem $\fIIIpIV$ with $\kappa(\boldsymbol{A}) = 10^6$ and $d_c = d_x = 10$ as a function of $a$. The notation conventions for the median and IQR, as well as the statistical methodology and the meaning of the significance symbols $\uparrow$ and $\downarrow$, follow \Cref{tab:fII}.}
\label{tab:fIIIpIV_kappa1e6_10_10}
\begin{tabular*}{\textwidth}{@{\extracolsep{\fill}}ccccc}
\hline
$a$ & & IGBD (ours) & CatCMA $\phantom{\downarrow}$ & ICatCMA $\phantom{\downarrow}$ \\ \hline
\multirow{3}{*}{0} & Success rate & \textcolor{blue}{0.95} & 0.00 $\downarrow$ & 0.00 $\downarrow$ \\
 & Total FEs & \textcolor{blue}{8.15E+5 (9.35E+5)} & 2.00E+6 (0.00E+0) $\uparrow$ & 2.00E+6 (0.00E+0) $\uparrow$ \\
 & Restarts & 0 (1) & 454 (6) $\uparrow$ & 8 (0) $\uparrow$ \\ \hline
\multirow{3}{*}{1} & Success rate & 0.85 & 0.00 $\downarrow$ & 0.00 $\phantom{\downarrow}$ \\
 & Total FEs & 6.25E+5 (5.93E+5) & 2.00E+6 (0.00E+0) $\phantom{\downarrow}$ & 2.00E+6 (0.00E+0) $\phantom{\downarrow}$ \\
 & Restarts & 0 (0) & 461 (5) $\uparrow$ & 8 (0) $\uparrow$ \\ \hline
\multirow{3}{*}{2} & Success rate & \textcolor{blue}{1.00} & 0.00 $\downarrow$ & 0.00 $\downarrow$ \\
 & Total FEs & \textcolor{blue}{4.17E+5 (4.87E+5)} & 2.00E+6 (0.00E+0) $\uparrow$ & 2.00E+6 (0.00E+0) $\uparrow$ \\
 & Restarts & 0 (0) & 468 (2) $\uparrow$ & 8 (1) $\uparrow$ \\ \hline
\multirow{3}{*}{4} & Success rate & \textcolor{blue}{1.00} & 0.00 $\downarrow$ & 0.00 $\downarrow$ \\
 & Total FEs & \textcolor{blue}{5.27E+5 (3.44E+5)} & 2.00E+6 (0.00E+0) $\uparrow$ & 2.00E+6 (0.00E+0) $\uparrow$ \\
 & Restarts & 0 (1) & 470 (4) $\uparrow$ & 6 (1) $\uparrow$ \\ \hline
\multirow{3}{*}{8} & Success rate & \textcolor{blue}{1.00} & 0.00 $\downarrow$ & 0.00 $\downarrow$ \\
 & Total FEs & \textcolor{blue}{4.52E+5 (5.29E+5)} & 2.00E+6 (0.00E+0) $\uparrow$ & 2.00E+6 (0.00E+0) $\uparrow$ \\
 & Restarts & 0 (0) & 436 (7) $\uparrow$ & 6 (1) $\uparrow$ \\ \hline
\multirow{3}{*}{16} & Success rate & \textcolor{blue}{1.00} & 0.00 $\downarrow$ & 0.00 $\downarrow$ \\
 & Total FEs & \textcolor{blue}{6.07E+5 (6.41E+5)} & 2.00E+6 (0.00E+0) $\uparrow$ & 2.00E+6 (0.00E+0) $\uparrow$ \\
 & Restarts & 0 (1) & 523 (10) $\uparrow$ & 9 (4) $\uparrow$ \\ \hline
\end{tabular*}
\end{table}
\end{document}